%% file: arxiv.tex
\documentclass{resources/hcrg/hcrg-nasa}
\issuer{Harvard Computational Robotics Group}
\edition{NeurIPS 2026}
\logofile{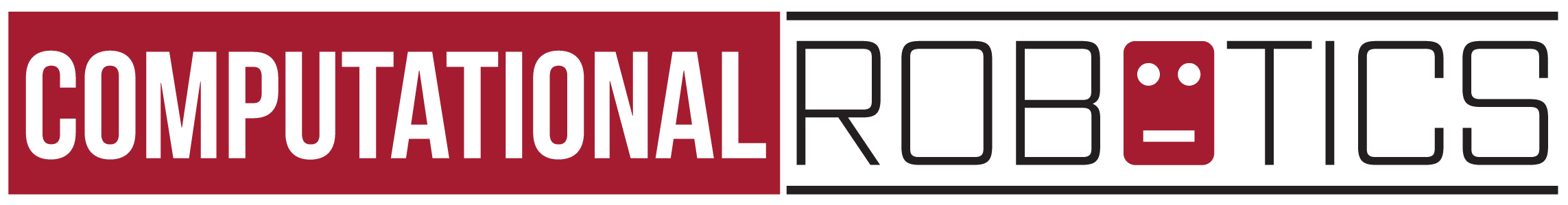}
\usepackage{etoolbox}
\patchcmd{\mymaketitle}{\noindent\includegraphics[width=\logowidth]{\logofilelist}\par}{}{}{\ClassWarning{hcrg-nasa}{wordmark patch did not apply}}

\usepackage[utf8]{inputenc}
\usepackage{url}
\usepackage{nicefrac}
\usepackage{wrapfig}
\usepackage{needspace}
\usepackage{algorithm}

\makeatletter
\renewcommand{\If}{\csname ALG@cmd@\ALG@L @If\endcsname}
\renewcommand{\For}{\csname ALG@cmd@\ALG@L @For\endcsname}
\renewcommand{\Return}{\csname ALG@cmd@\ALG@L @Return\endcsname}
\def\theHALG@line{\theHalgorithm.\arabic{ALG@line}}
\makeatother
\newtheorem{theorem}{Theorem}

\input{arxiv_source/shortcuts}

\title{\name: Cross-Domain Co-Training of Generative Robot Policies via Best-Effort Adaptation}
\author[1]{Antong Zhang}
\author[2]{Han Qi}
\author[2]{Heng Yang}
\affiliation[1]{Department of Computer Science, Brown University}
\affiliation[2]{School of Engineering and Applied Sciences, Harvard University}
\contribution[]{\texttt{antong\_zhang@brown.edu}\quad\texttt{hqi@g.harvard.edu}\quad\texttt{hankyang@seas.harvard.edu}}
\abstract{\input{arxiv_source/abstract.tex}}
\website{https://computationalrobotics.seas.harvard.edu/BEACON/}{computationalrobotics.seas.harvard.edu/BEACON/}

\begin{document}
\noindent\IfFileExists{resources/hcrg/hcrg-wordmark.png}{\includegraphics[width=\logowidth]{resources/hcrg/hcrg-wordmark.png}}{\mbox{}}\par
\maketitle

\input{arxiv_source/sections/introduction.tex}
\input{arxiv_source/sections/related-work.tex}
\input{arxiv_source/sections/bea.tex}
\input{arxiv_source/sections/beacon.tex}

\input{arxiv_source/sections/experiments.tex}

\input{arxiv_source/sections/conclusion.tex}

\clearpage
\section*{Acknowledgments}
We acknowledge funding from the Office of Naval Research grant N00014-25-1-2322.

\bibliographystyle{unsrtnat}
\bibliography{arxiv_source/references}

\clearpage
\appendix
\setcounter{figure}{0}
\renewcommand{\thefigure}{A\arabic{figure}}
\renewcommand{\theHfigure}{appendix.\arabic{figure}}
\setcounter{algorithm}{0}
\renewcommand{\thealgorithm}{A\arabic{algorithm}}
\renewcommand{\theHalgorithm}{appendix.alg.\arabic{algorithm}}
\input{arxiv_source/sections/appendix/A.tex}

\input{arxiv_source/sections/appendix/B.tex}
\input{arxiv_source/sections/appendix/C.tex}
\input{arxiv_source/sections/appendix/D.tex}
\input{arxiv_source/sections/appendix/E.tex}
\input{arxiv_source/sections/appendix/F.tex}
\input{arxiv_source/sections/appendix/G.tex}

\end{document}

%% file: arxiv_source/shortcuts.tex
\newcommand{\name}{\textsc{BEACON}}



%% file: arxiv_source/abstract.tex
  
  We introduce \name{}---Best-Effort Adaptation for Cross-Domain Co-Training---a theory-guided framework for training generative robot policies with abundant source demonstrations and limited target demonstrations. \name{} casts cross-domain co-training as a discrepancy-aware importance-reweighting problem, jointly learning a diffusion-based visuomotor policy and per-sample source weights through an objective inspired by the Best-Effort Adaptation (BEA) bound. To make this approach practical for high-dimensional sequence policies, we develop scalable instance-level discrepancy proxies, stochastic alternating updates for policy and weights, and a multi-source extension that balances heterogeneous source domains.
  Across sim-to-sim, sim-to-real, and multi-source manipulation settings, \name{} improves robustness and data efficiency over target-only, fixed-ratio co-training, and feature-alignment baselines. Without an explicit alignment objective, \name{} also brings source and target representations closer together as an implicit result of discrepancy-aware co-training. 
  

%% file: arxiv_source/sections/introduction.tex

\section{Introduction}

Behavior cloning (BC) has become a prominent route to learning generalizable robot manipulation policies because it reduces policy training to supervised imitation from expert demonstrations~\citep{pomerleau1988alvinn,billard2008robot,qi26icra-compose,qi26iral-gpc}. Recent sequence models, especially diffusion-based policies, further strengthen this paradigm by modeling temporally extended and multimodal action distributions~\citep{zhao2023learning,chi2023diffusion}. Despite this progress, policies trained in one domain often degrade in another even when task semantics are unchanged. Source and target domains can differ in appearance and camera viewpoint, while simulation-to-real transfer also involves differences in contact and dynamics~\citep{rusu2017sim,tobin2017domain,peng2018sim}. Directly reusing source demonstrations can therefore be unreliable, while collecting large target-domain datasets is expensive and difficult to scale~\citep{ebert2021bridge,brohan2022rt1,mandlekar2023mimicgen}. This raises a central question: \emph{how can we leverage abundant but mismatched source demonstrations while using a minimal amount of target data?}

Classical methods use \emph{domain randomization} to increase source diversity~\citep{tobin2017domain,sadeghi2017cad2rl,peng2018sim}, while \emph{domain adaptation} approaches explicitly align source and target representations to promote invariance to domain shifts~\citep{bousmalis2017unsupervised,ganin2016domain,long2015learning,courty2017optimal}. More recent approaches show that \emph{co-training} on source and target demonstrations can be effective when target data is scarce~\citep{maddukuri2025sim,cheng2025generalizable}. However, these co-training approaches optimize empirical objectives through fixed mixing rules or auxiliary alignment losses. We focus instead on which source demonstrations should supervise the target policy. The question then follows. \emph{How can target-domain generalization bounds guide the design of training objectives that adaptively weight source demonstrations?}

\begin{figure}
  \centering
  \includegraphics[width=\textwidth]{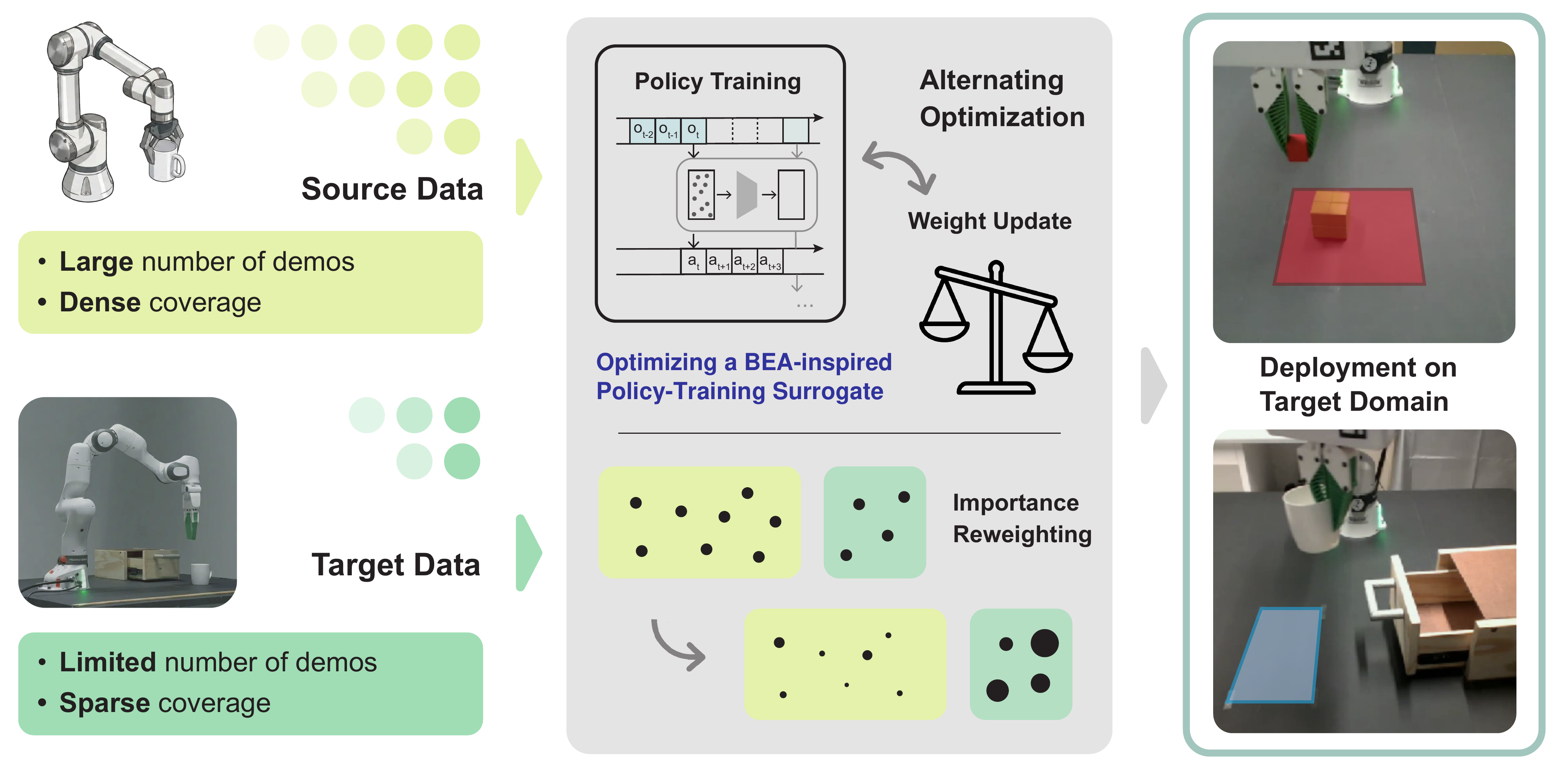}
  \vspace{-6mm}
  \caption{{The \name{} framework.} Given a large amount of (multi-)source demonstrations and a small amount of target demonstrations, 
  \name{} jointly learns a diffusion-based visuomotor policy and a set of \emph{per-sample weights} using a \emph{discrepancy-aware objective inspired by BEA}.
  }
  \label{fig:framework}
  \vspace{-4mm}
\end{figure}

\textbf{Contribution.} We take a theory-guided perspective and revisit the classical literature in domain adaptation, where target-domain generalization bounds guide the design of learning objectives. In particular, we build on the recent Best-Effort Adaptation (BEA) framework~\citep{awasthi2023best}, and adapt its source-weighting principle to cross-domain diffusion-policy training for robot learning.

Concretely, we cast cross-domain co-training as an \emph{importance-reweighting} problem with abundant source data and limited target data. The core idea is to reweight source samples using an instance-level discrepancy proxy to favor those more relevant to the target. This yields a \emph{joint optimization} over \emph{policy parameters} and \emph{source weights}, guided by BEA's empirical-loss, discrepancy, and regularization terms. Applying this BEA-inspired formulation to modern manipulation tasks, however, requires practical adaptations for high-dimensional diffusion policies. We develop scalable discrepancy proxies and stochastic alternating updates tailored to sequence prediction~\citep{cortes2019adaptation,zhang2020localized,ganin2016domain,cover1967nearest}. We also extend the method to multi-source settings motivated by realistic data pipelines, where demonstrations may come from different simulators, real-world scene configurations, or prior datasets, combining domain-level weights with weights on individual training windows. We call the method \name{} (Best-Effort Adaptation for Cross-Domain Co-Training, Figure~\ref{fig:framework}) and evaluate it across sim-to-sim, sim-to-real, and multi-source manipulation, where it improves transfer over fixed-mixture co-training and feature-alignment baselines. Our analysis further shows that \name{} brings source and target representations closer together without an explicit feature-alignment loss.

\textbf{Outline.} The rest of the paper is organized as follows. Section \ref{sec:related_work} reviews related work. Section \ref{sec:bea} introduces the theoretical foundations of BEA. Section \ref{sec:beacon} describes our \name{} method. Section \ref{sec:experiments} presents the experiments and analysis. Section \ref{sec:conclusion} concludes the paper.

%% file: arxiv_source/sections/related-work.tex

\section{Related work}
\label{sec:related_work}

\textbf{Behavior cloning for robotic manipulation.}
Behavior cloning reduces policy learning to supervised prediction from demonstration data~\citep{pomerleau1988alvinn,billard2008robot,qi25iclr-control}. Recent progress has come from both scaling real-robot datasets and improving sequence models for visuomotor control. Large multi-task and multi-domain robot datasets such as BridgeData and RT-1 show that diversity and scale materially improve generalization~\citep{ebert2021bridge,brohan2022rt1}. Simulation benchmarks and datasets such as LIBERO and RoboCasa further expand the space of procedurally generated tasks and environments~\citep{liu2023libero,robocasa2024}. In parallel, ACT predicts action chunks for precision manipulation, while Diffusion Policy models high-dimensional, multimodal action distributions~\citep{zhao2023learning,chi2023diffusion}. Recent analyses of transformer-based vision--language--action models report that action-aligned latent geometry emerges during training and tracks task performance~\citep{qi2026actionaligned}. At the same time, these successes accentuate the data bottleneck: collecting high-quality real demonstrations remains slow and expensive~\citep{maddukuri2025sim}. Systems such as MimicGen partially alleviate this problem by synthesizing large simulation datasets from a small set of human demonstrations~\citep{mandlekar2023mimicgen}. We build on this foundation of diffusion-based BC, focusing specifically on how to effectively leverage large-scale cross-domain source demonstrations (e.g., from simulation) alongside a limited dataset of target domain examples.

\textbf{Sim-to-Real adaptation and Sim-and-Real co-training.}
Existing work addresses sim-to-real transfer primarily by reducing mismatch between simulated and real environments. Classical approaches such as domain randomization improve robustness by exposing policies to broader variation in simulation~\citep{tobin2017domain,sadeghi2017cad2rl,peng2018sim}, while domain adaptation methods explicitly align the two domains at the pixel or feature level~\citep{bousmalis2017unsupervised,hoffman2018cycada,ganin2016domain,long2015learning,courty2017optimal}. Recent sim-and-real co-training methods further show that jointly training a single policy on mixed simulated and real demonstrations can be highly effective~\citep{maddukuri2025sim,lei2026mechanistic}. Cheng et al.~\citep{cheng2025generalizable} augment co-training with task-aware Unbalanced Optimal Transport (UOT), using proprioceptive guidance and temporally aligned sampling to learn transferable representations. Its relaxed mass constraints accommodate unmatched portions of the two distributions. Although alignment-based co-training methods such as UOT show improvements, aligning feature distributions does not by itself ensure that the aligned features remain relevant to action under physical domain shifts. Their effectiveness also depends on which representations are aligned and how similarity is measured.

Our approach addresses the complementary question of which source examples should supervise the target policy. We learn their contributions to the behavior-cloning loss through importance reweighting, using policy features to estimate source--target mismatch without adding a feature-matching loss. This weighting formulation also extends naturally to multiple source domains.

\textbf{Data weighting and curation for imitation learning.}
Recent methods select useful robot data at different levels. ReMix learns domain-level mixtures for robust imitation pretraining through distributionally robust optimization~\citep{hejna2025remix}. CUPID uses evaluation rollouts and influence estimates to rank demonstrations by their contribution to a policy's expected return~\citep{agia2025cupid}. \name{} instead learns target-specific weights on source observation windows jointly with the policy, using the available source and target demonstrations. These approaches provide complementary ways to decide how training data should influence policy learning.

\textbf{Sample reweighting and multi-source domain adaptation.}
Classical work on distribution shift and domain adaptation characterized domain mismatch and motivated learning algorithms that account for differences between source and target distributions~\citep{kifer2004detecting,ben2010theory}. Subsequent work developed task-aware discrepancy notions that depend on the loss and hypothesis class, yielding bounds and algorithms for regression, sample-bias correction, and generalized discrepancy minimization~\citep{mansour2009bounds,cortes2014domain,cortes2019adaptation}. Classical importance-weighting approaches estimate weights from distribution mismatch before fitting the predictor~\citep{huang2007correcting,sugiyama2007direct}, with learning bounds characterizing the effect of those weights~\citep{cortes2010learning}. Multi-source domain adaptation studies the setting where labeled data are available from multiple source distributions~\citep{mansour2009domain,zhao2024more}. Prior work shows that naively combining sources can underperform the best individual source because domain shift exists not only between each source and the target, but also among sources~\citep{zhao2018adversarial,wen2020domain,zhao2024more}. Existing approaches use methods like source-specific alignment, domain aggregation, and learned source weighting to control negative transfer~\citep{zhao2018adversarial,peng2019moment,wen2020domain}.

\name{} follows this discrepancy-based source-weighting direction, particularly the recent Best-Effort Adaptation framework~\citep{awasthi2023best}, which explicitly assumes limited labeled target-domain data. Our key contribution is to adapt BEA to modern diffusion-based robot policy learning with practical enhancements for high-dimensional sequence prediction.

%% file: arxiv_source/sections/bea.tex

\section{Problem setting and theoretical foundations}
\label{sec:bea}

We study supervised learning over an input--output space $\mathcal{X}\times\mathcal{Y}$ with a
source distribution $Q$ and a
target distribution $P$. We observe a labeled
source dataset $D_{\mathrm{src}}=\{(x_i,y_i)\}_{i=1}^m \sim Q^m$
and a labeled target dataset
$D_{\mathrm{tgt}}=\{(x_j,y_j)\}_{j=m+1}^{m+n} \sim P^n,$
with typically $m\gg n$. Let $S=\{1,\dots,m\}$ and $T=\{m+1,\dots,m+n\}$ denote the source and
target index sets. The goal is to learn a hypothesis $h\in\mathcal{H}$ that minimizes the target
population loss
\begin{equation}\label{eq:target-risk}
L(P,h)=\mathbb{E}_{(x,y)\sim P}\big[\ell(h(x),y)\big],
\end{equation}
for a bounded loss $\ell:\mathcal{Y}\times\mathcal{Y}\to[0,1]$. BEA \citep{awasthi2023best}
asks how to use abundant but potentially mismatched source data to improve performance on $P$,
rather than naively mixing all samples equally.

\textbf{Source reweighting.} The key idea is to assign a nonnegative weight to each training example and learn both the predictor
and the weights jointly. Concretely, we introduce a weight vector
$\mathbf{q}=(q_1,\dots,q_{m+n})\in\mathbb{R}_+^{m+n}$,
where each $q_i\ge 0$ controls the contribution of sample $(x_i,y_i)$. Intuitively, source examples
that are more relevant to the target domain should receive larger weight, while mismatched source
examples should be down-weighted. 

\textbf{Discrepancy measure.} To formalize when source data is helpful, BEA uses a discrepancy measure that quantifies domain
mismatch in a way that is aligned with both the loss and the hypothesis class. For two labeled
distributions $P$ and $Q$, the one-sided labeled discrepancy is
\begin{equation}\label{eq:discrepancy-measure}
\mathrm{dis}(P,Q)
:=
\sup_{h\in\mathcal H}
\left(
\mathbb{E}_{(x,y)\sim P}[\ell(h(x),y)]
-
\mathbb{E}_{(x,y)\sim Q}[\ell(h(x),y)]
\right).
\end{equation}
Unlike a generic divergence between distributions, this quantity only measures differences that matter
for predictors in $\mathcal H$ under the loss $\ell$. 
At the sample level, BEA also introduces a discrepancy between the learned weighting
$\mathbf{q}$ and a reference weighting $\mathbf{p}_0$. A natural choice for $\mathbf{p}_0$ is the
target-only empirical distribution, $(\mathbf{p}_0)_i = \frac{1}{n}$ for $i\in T$ and $(\mathbf{p}_0)_i = 0$ for $i\in S$.
This choice reflects the idealized baseline in which, absent any trustworthy source information, one
would train only on the target sample. Given this reference, the weighted sample discrepancy is
\begin{equation}
\label{eq:weighted-sample-discrepancy}
\mathrm{dis}(\mathbf{q},\mathbf{p}_0)
:=
\sup_{h\in\mathcal H}
\sum_{i=1}^{m+n}\big(q_i-(\mathbf{p}_0)_i\big)\,\ell(h(x_i),y_i).
\end{equation}
This one-sided quantity measures an excess of the $\mathbf{q}$-weighted empirical loss over the target empirical loss, uniformly over $\mathcal H$.

\textbf{Generalization bound.} These quantities enter a discrepancy-based generalization bound due to \citep[Corollary 4]{awasthi2023best},
which makes precise the trade-off between fitting the weighted training data and keeping the effective
source contribution close to the target domain. 


\begin{theorem}[Discrepancy-based generalization bound \citep{awasthi2023best}]
\label{thm:discrepancy-generalization}
For any $\delta>0$, with probability at least $1-\delta$ over the choice of a source sample
$D_{\mathrm{src}}=\{(x_i,y_i)\}_{i=1}^m \sim Q^m$ and a target sample
$D_{\mathrm{tgt}}=\{(x_j,y_j)\}_{j=m+1}^{m+n} \sim P^n$, the following holds simultaneously for all
$h\in\mathcal H$ and all weight vectors $\mathbf{q}\in[0,1]^{m+n}$ satisfying
$\|\mathbf{q}-\mathbf{p}_0\|_1<1$:
\begin{align}
L(P,h)
\le\;&
\sum_{i=1}^{m+n} q_i\,\ell(h(x_i),y_i)
+ q_{\mathrm{src}}\,\mathrm{dis}(P,Q)
+ \mathrm{dis}(\mathbf{q},\mathbf{p}_0)
+ 2\mathfrak{R}_{\mathbf{q}}(\ell\circ\mathcal H) + 6\|\mathbf{q}-\mathbf{p}_0\|_1
\nonumber\\
&\quad
+ \Big(\|\mathbf{q}\|_2+2\|\mathbf{q}-\mathbf{p}_0\|_1\Big)
\left[
\sqrt{\log\log_2\!\frac{2}{1-\|\mathbf{q}-\mathbf{p}_0\|_1}}
+
\sqrt{\frac{\log(2/\delta)}{2}}
\right],
\label{eq:bea-bound}
\end{align}
where $q_{\mathrm{src}}:=\sum_{i\in S} q_i$, $\mathrm{dis}(P,Q)$ is the labeled discrepancy \eqref{eq:discrepancy-measure} between
the source and target distributions, $\mathrm{dis}(\mathbf{q},\mathbf{p}_0)$ is the weighted sample
discrepancy \eqref{eq:weighted-sample-discrepancy} between $\mathbf{q}$ and the reference weights $\mathbf{p}_0$, and
$\mathfrak{R}_{\mathbf{q}}(\ell\circ\mathcal H)$ is the $\mathbf{q}$-weighted Rademacher complexity.
\end{theorem}

The first term in Theorem~\ref{thm:discrepancy-generalization} is the weighted empirical loss, so it encourages selecting examples that improve prediction.
The second term shows that source data is only useful to the extent that it remains compatible with
the target domain: large source mass is penalized when $\mathrm{dis}(P,Q)$ is large. The third term
penalizes choices of $\mathbf{q}$ whose induced empirical objective deviates too far from the
target-only reference $\mathbf{p}_0$. Finally, the remaining terms control complexity and suppress
degenerate solutions that place too much mass on only a few examples. 


\textbf{The SBEST objective in BEA.} Assuming the complexity term $\mathfrak{R}_{\mathbf{q}}(\ell\circ\mathcal H)$ can be upper bounded by a norm penalty on $h$, \citet{awasthi2023best} derive the SBEST learning objective from the BEA bound. Defining $d_i := \mathrm{dis}(P,Q)\,\mathbf{1}[i\in S]$ folds the source-discrepancy term into the weighted loss. Since $\ell\in[0,1]$, the weighted sample discrepancy is upper bounded by $\|\mathbf{q}-\mathbf{p}_0\|_1$.\footnote{BEA~\citep{awasthi2023best} also presents a variant that retains the weighted sample discrepancy term explicitly, yielding a more complex optimization problem.} Absorbing the bound constants and concentration terms into regularization coefficients yields SBEST:
\begin{equation}
\min_{h\in\mathcal H,\;\mathbf{q}\in[0,1]^{m+n}}
\sum_{i=1}^{m+n} q_i\big(\ell(h(x_i),y_i)+d_i\big)
+ \lambda_\infty \|\mathbf{q}\|_\infty \|h\|^2
+ \lambda_1\|\mathbf{q}-\mathbf{p}_0\|_1
+ \lambda_2\|\mathbf{q}\|_2^2,
\label{eq:sbest-objective}
\end{equation}
where $\lambda_\infty, \lambda_1, \lambda_2 \ge 0$ are tunable coefficients. In BEA's setting, this bound-derived objective jointly learns a predictor and sample weights that fit the weighted data, limit the influence of discrepant source data, and remain anchored to the target empirical distribution. Section~\ref{sec:beacon} describes our adaptation of this objective to diffusion-policy training.

%% file: arxiv_source/sections/beacon.tex
\section{Best-Effort Adaptation for cross-domain policy co-training}
\label{sec:beacon}


We now instantiate the BEA framework for learning
visuomotor robot manipulation policies.
We consider behavior cloning from demonstration trajectories collected in a source domain and a target
domain. At each decision time $t$, the policy receives a window $x_t$ of recent visual observations and proprioceptive features, and the desired output is an action chunk $y_t$. Each such window forms one training example $(x_i,y_i)$; we use $m,n$ for source and target window counts, and each window has its own weight.
We learn a policy $\pi_\theta$ with the architecture of a Diffusion Policy~\cite{chi2023diffusion}, trained by behavior cloning.
Accordingly, the hypothesis $h\in\mathcal H$ is replaced by $\pi_\theta$,
and the generic loss $\ell(h(x_i),y_i)$ is instantiated as the per-example diffusion behavior cloning
loss
$
\ell_{\mathrm{BC}}(\theta; x_i,y_i).
$ We adapt the BEA objective \eqref{eq:sbest-objective} to policy learning as follows:
\begin{equation}
\min_{\theta,\;\mathbf q\in\mathcal C_b}
\sum_{i=1}^{m+n} q_i\big(\ell_{\mathrm{BC}}(\theta; x_i,y_i)+ \lambda_d d_i\big)
+\gamma \|\mathbf q\|_\infty R(\theta)
+\lambda_1\|\mathbf q-\mathbf r\|_1
+\lambda_2\|\mathbf q\|_2^2,
\label{eq:beacon-objective}
\end{equation}
where $R(\theta)$ is a policy regularizer and $\mathbf r$ is a uniform target-only reference, with $r_i$ equal to one over the number of target demonstrations for target windows and $0$ for source windows. The feasible set $\mathcal C_b=\{\mathbf q:\underline q_i\le q_i\le q_{\max},\ \sum_iq_i=b\}$ uses a target floor $\underline q_i=q_T^{\min}$ for $i\in T$, zero source floor, and total budget $b=n+\alpha m$. Target weights remain adaptive, so $\alpha$ controls the total budget rather than fixing source mass. The default discrepancy proxy $d_i$ is nonnegative; we set $d_i=0$ for target windows and tune $\lambda_d\ge0$ to control the source penalty.


\textbf{Challenges in applying BEA to visuomotor policy learning.}
It is nontrivial to carry the original BEA framework over to visuomotor policy learning. First, the original BEA analysis is developed for bounded supervised losses and hypothesis classes
whose complexity can be controlled by relatively simple norm-based quantities. Diffusion Policies,
however, are high-capacity sequence models trained with denoising-style objectives, and their
optimization and generalization behavior are not well captured by the settings
in which BEA was developed and evaluated.
Second, the original BEA framework solves \eqref{eq:sbest-objective} through full-batch
alternating minimization over $h$ and $\mathbf{q}$. In image-conditioned robot policy learning,
however, even a single loss evaluation requires a forward pass through a large diffusion model on
high-dimensional observation histories. At the scales relevant to our setting, both policy updates and
weight updates must therefore be carried out stochastically over minibatches rather than through
full-batch optimization.
Third, the discrepancy term $d_i$ in BEA relies on optimization over relatively simple hypothesis classes with full
access to labeled data. In our setting, the inputs are RGB observation windows, the outputs are
action chunks, and the predictor is a deep generative policy. Solving the original discrepancy
estimation problem directly in this space would be computationally prohibitive.

These mismatches motivate a practical adaptation of BEA for high-dimensional
robot policy learning that preserves the structure of the bound-informed objective while replacing the
parts of the original framework that do not scale to diffusion-based visuomotor control. Relative to \eqref{eq:sbest-objective}, \name{} replaces the bounded loss with the diffusion behavior-cloning loss, the domain-level discrepancy $\mathrm{dis}(P,Q)$ with an instance-level proxy $d_i$, the norm-based capacity term with weight decay scaled by $\|\mathbf q\|_\infty$, BEA's admissible weights with the feasible set $\mathcal C_b$, and full-batch alternating minimization with stochastic alternating updates.

\subsection{The \name{} algorithm}

\textbf{Complexity regularization via optimizer-compatible weight decay.}
In the original BEA objective \eqref{eq:sbest-objective}, the capacity control term has the form
$\lambda_\infty \|\mathbf q\|_\infty \|h\|^2$. For diffusion policies, we replace this by
$
\gamma \|\mathbf q\|_\infty R(\theta)$ with $R(\theta):=\frac{1}{2}\|\theta\|_2^2$,
where $\gamma$ sets the strength of weight decay. In the policy update, we scale the AdamW decay by $\|\mathbf q\|_\infty$, preserving the intuition that larger maximal sample weights should incur stronger regularization.

\textbf{Instance-level discrepancy instead of domain-level discrepancy.}
A direct instantiation of BEA would assign the same discrepancy penalty
$d_i=\mathrm{dis}(P,Q)\mathbf 1[i\in S]$ to every source sample. In robot manipulation, however,
source demonstrations are highly heterogeneous: even within one simulator, some trajectories are
visually and behaviorally close to the target domain, while others are clear outliers. We therefore
replace the domain-level penalty by an instance-level discrepancy score $d_i$.

Our default proxy computes $d_i$ in the policy's observation-encoder embedding space using
$k$-nearest neighbors. Specifically, for each source sample $x_i$, we measure the average Euclidean
distance from its embedding to its $k$ nearest target embeddings, normalized by the mean within-target
$k$-NN distance. Since the encoder is trained by behavior cloning on action chunks, $d_i$ is not a pixel distance. The discrepancy scores are refreshed before each weight-update phase and held fixed during the subsequent updates.

We also consider two alternatives as ablations. The first is a localized labeled discrepancy estimator
closer in spirit to BEA, which maximizes empirical discrepancy over a local policy class around a
target-trained reference model. The second is a domain-classifier-based estimator that fits a binary
discriminator on the encoder features and uses its per-source posterior as an instance-level mismatch signal. Implementation details are provided in the appendix. These variants help understand the role of discrepancy estimation, while the default $k$-NN estimator provides window-level scores without a separate reference policy or inner optimization.

\textbf{Stochastic alternating optimization over policy and weights.}
We optimize \eqref{eq:beacon-objective}
by alternating stochastic updates of $\theta$ and $\mathbf q$. For a minibatch $B\subseteq \{1,\dots,m+n\}$,
we first compute the per-example loss $\ell_i(\theta):=\ell_{\mathrm{BC}}(\theta;x_i,y_i)$ and then update
the weights through projected subgradient:
\begin{equation}
g_i
=
\ell_i(\theta)
+\lambda_d d_i
+\gamma R(\theta)\,
v_i
+\lambda_1\,\mathrm{sign}(q_i-r_i)
+2\lambda_2 q_i,
\qquad i\in B,
\label{eq:q-grad}
\end{equation}
where $v_i$ splits the capacity subgradient equally among minibatch entries attaining $\|\mathbf q\|_\infty$ and is zero otherwise. We update $q_i\leftarrow\mathrm{clip}_{[\underline q_i,q_{\max}]}(q_i-\eta_qg_i)$ and project onto $\mathcal C_b$ after the full weight sweep. The policy is then updated with AdamW on the weighted minibatch loss $\frac{1}{|B|}\sum_{i\in B}q_i\ell_i(\theta)$.
This stochastic alternation is the practical analogue of BEA's joint optimization.

As an ablation, we also implement a convex inner update, closer to BEA's alternating minimization, that solves
the weight subproblem at fixed $\theta$ and auxiliary variable $\|\mathbf q\|_\infty$. We found that the projected-subgradient update
is simpler and better suited to large-scale training. More details are provided in the appendix.

\textbf{Extension to multi-source co-training.}
In many realistic pipelines, the source data does not come from a single homogeneous domain, but
from multiple simulators, scene configurations, or prior datasets. Pooling them into one
source distribution can amplify negative transfer because mismatch exists both between each source and
the target and among the source domains themselves.

To address this, we factor the source weight into two levels: a domain-level weight
$\mathbf w\in\Delta^{K-1}$ over $K$ source domains with index sets $S_s$, and a within-domain nonnegative weight
$\widetilde{\mathbf q}$. If sample $i\in S$ comes from source domain $s(i)$, we define
$
q_i = w_{s(i)}\widetilde q_i, i\in S$,
and
$q_j=\widetilde q_j, j\in T$.
This yields a multi-source \name{} objective in which $\mathbf w$ controls which source domains are
useful globally, while $\widetilde{\mathbf q}$ suppresses mismatched training windows within each source.
The multi-source objective evaluates \eqref{eq:beacon-objective} on these composite weights and adds $\rho_1\|\mathbf w-\mathbf w_0\|_1+\rho_2\|\mathbf w\|_2^2$, with $\mathbf w_0=\mathbf 1/K$. Training alternates updates of the domain weights, within-domain weights, and policy. Full method details are provided in the appendix.

%% file: arxiv_source/sections/experiments.tex
\section{Experiments}
\label{sec:experiments}

The experiments are designed to substantiate five claims about \name{}. \textbf{(i)} Learned source weighting improves cross-domain policy performance relative to fixed-mixture co-training and feature-alignment baselines in the evaluated sim-to-sim and sim-to-real settings. \textbf{(ii)} \name{} learns policies that generalize to out-of-distribution task instances. \textbf{(iii)} Multi-source \name{} effectively combines heterogeneous source domains. \textbf{(iv)} Closer feature alignment does not necessarily yield better transfer. \textbf{(v)} \name{} shows an upward trend as the number of target demonstrations increases.


\begin{figure}
    \centering
    \includegraphics[width=\textwidth]{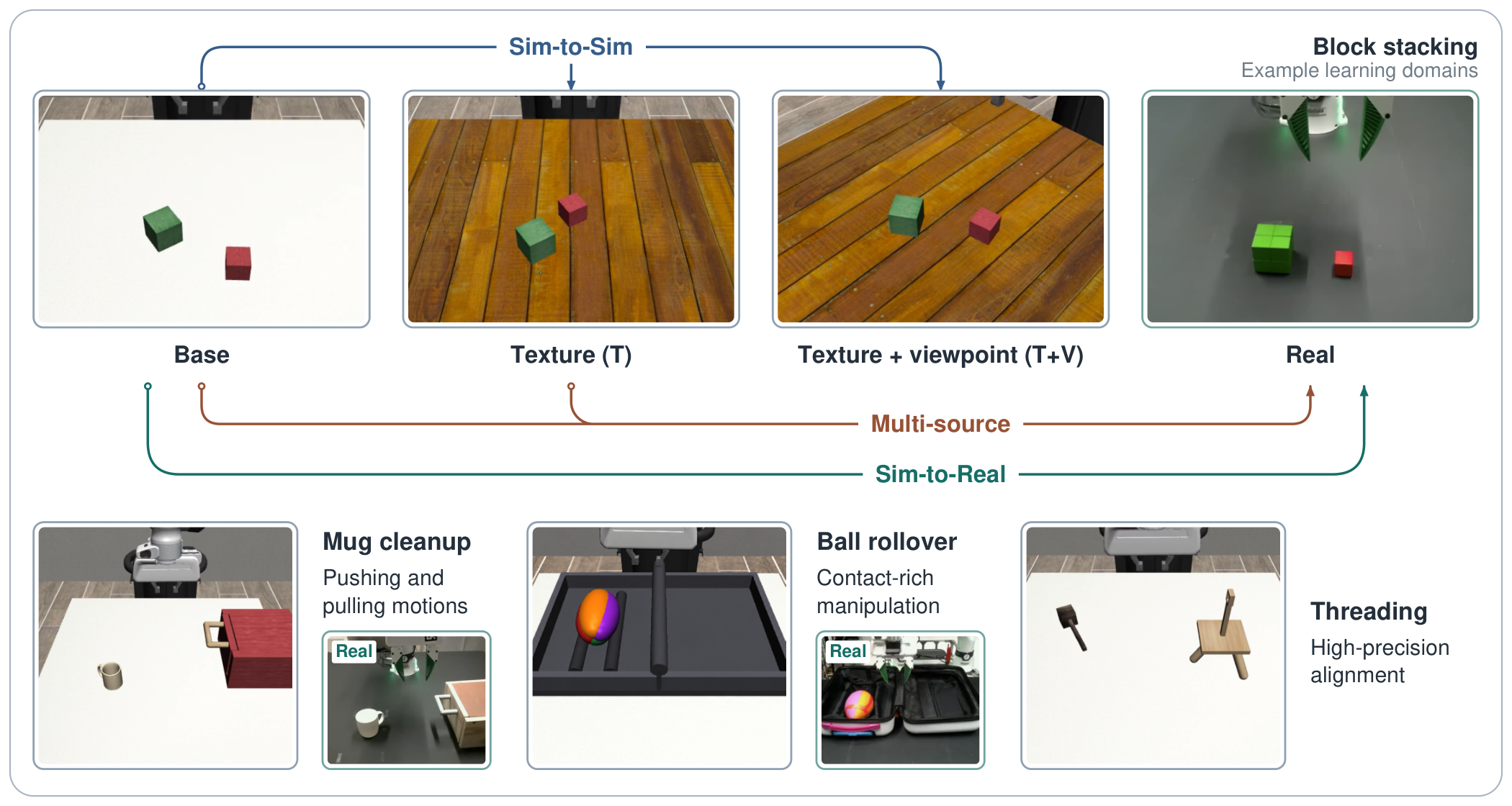}
    \vspace{-3mm}
    \caption{{Experimental setup.} Top shows block-stacking domains and transfer settings. Target-T changes the texture, and Target-T+V changes the texture and camera viewpoint relative to the base simulation. The real domain has both visual and physical gaps. Bottom shows demonstrations for other tasks that involve different motion patterns.}
    \label{fig:tasks}
    \vspace{-3mm}
\end{figure}

\subsection{Experimental setup}

We study image-based visuomotor manipulation on a Franka Emika Panda robot arm across four tasks. Our sim-to-sim benchmarks are \emph{block stacking}, \emph{mug cleanup}, and \emph{threading}. Sim-to-real experiments use stacking, mug cleanup, and the contact-rich task of \emph{ball rollover}, in which the robot rolls a football to a suitcase's central ridge and levers it into the opposite compartment (Figure~\ref{fig:tasks}). Tasks are evaluated in two cross-domain regimes: a controlled \emph{sim-to-sim} setting, in which the source and target are both simulated but differ in their visual observation distributions, and a \emph{sim-to-real} setting, in which the target is collected on the physical robot with both visual and physical gaps.

\textbf{Visual gaps.} The policy receives RGB observations resized to $96 \times 96$ together with end-effector proprioception, and the visual stream is encoded by a ResNet-18 backbone shared across domains. In simulation we instantiate two visual gaps between source and target: a \emph{texture} gap, in which scene and object textures differ while the camera pose is held fixed, and a \emph{texture + viewpoint} gap, which additionally perturbs the camera extrinsics. The sim-to-real gap is the natural mismatch between rendered simulation imagery and real RGB frames captured by a fixed camera observing the physical workspace, which subsumes background and texture differences.

\textbf{Learning domains.} All simulated demonstrations are generated in robosuite~\citep{zhu2020robosuite} via MimicGen~\citep{mandlekar2023mimicgen}, which expands a small set of human seed demonstrations into a large source dataset. For every task we use a source set of 500 demonstrations. The target set has 20 demonstrations in the sim-to-sim and 40 in the sim-to-real tasks. The environment with \emph{texture gap} is used as the second source domain in the multi-source setting, where each source constitutes 50\% of the total source data.


\textbf{Tasks.} For \emph{block stacking}, both cubes are independently randomized over a square region of the workspace. For \emph{mug cleanup}, the mug's position and orientation are randomized over a rectangular region. For \emph{threading}, the thread's position and orientation are randomized over a rectangular region. For \emph{ball rollover}, the ball starts at a randomized position on one side of the suitcase. To probe extrapolation, we additionally run an out-of-distribution (OOD) evaluation on \emph{block stacking}, \emph{mug cleanup}, and \emph{ball rollover}: target demonstrations are collected with objects randomized over designated regions, while evaluation initial states are sampled exclusively from areas different from the target demonstrations. More details on task objectives and evaluation criteria are in the appendix.

\textbf{Baselines.} All methods use the same Diffusion Policy~\citep{chi2023diffusion} backbone and observation encoder. \emph{Source-Only} trains on $\mathcal{D}_\mathrm{src}$ alone, and \emph{Target-Only} trains on $\mathcal{D}_\mathrm{tgt}$ alone. The remaining baselines use both datasets. \emph{Co-Training} performs joint training on $\mathcal{D}_\mathrm{src} \cup \mathcal{D}_\mathrm{tgt}$ with a fixed source--target mixing ratio~\citep{maddukuri2025sim}. \emph{Tuned Uniform} assigns one common weight to all source losses, setting $q_i=c$ for every source window $i\in S$ instead of adapting $q_i$ individually. We select this weight on block stacking and reuse it across all sim-to-sim tasks and gaps (Appendix~\ref{app:hyperparam}). \emph{MMD}~\citep{long2015learning} augments co-training with a maximum mean discrepancy feature-alignment loss between source and target encoder activations. \emph{UOT}~\citep{cheng2025generalizable} replaces the alignment loss with unbalanced optimal transport and temporally aligned sampling.

\FloatBarrier
\subsection{Generalization results}

\begin{table}[htbp]
\caption{{Sim-to-sim mean task scores.} T denotes the texture gap and T+V the texture + viewpoint gap. Stacking reports relaxed (R) and strict (S) success as defined in Appendix~\ref{app:tasks}; mug cleanup includes half credit for partial completion. Each Average is the equal-weight mean over stacking (R), mug cleanup, and threading.}
\label{tab:sim2sim}
\centering
\small
\begin{tabular*}{\textwidth}{@{\extracolsep{\fill}}lcccccccccc@{}}
\toprule
 & \multicolumn{4}{c}{\textbf{Block Stacking}} & \multicolumn{2}{c}{\textbf{Mug Cleanup}} & \multicolumn{2}{c}{\textbf{Threading}} & \multicolumn{2}{c}{\textbf{Average}} \\
\cmidrule(lr){2-5} \cmidrule(lr){6-7} \cmidrule(lr){8-9}
\cmidrule(lr){2-3} \cmidrule(lr){4-5}
Method & T (R) & T (S) & T+V (R) & T+V (S) & T & T+V & T & T+V & T & T+V \\
\midrule
Source-Only           & 0.00 & 0.00 & 0.00 & 0.00 & 0.00 & 0.00 & 0.00 & 0.00 & 0.00 & 0.00 \\
Target-Only           & 0.16 & 0.10 & 0.23 & 0.20 & 0.16 & 0.12 & 0.15 & 0.16 & 0.16 & 0.17 \\
\addlinespace[2pt]
\cmidrule(lr){1-11}
\addlinespace[2pt]
Co-Training           & 0.33 & 0.25 & 0.26 & 0.16 & 0.29 & 0.15 & 0.33 & 0.14 & 0.32 & 0.18 \\
Tuned Uniform         & 0.62 & 0.45 & 0.47 & 0.37 & 0.34 & 0.17 & 0.38 & 0.18 & 0.45 & 0.27 \\
MMD                   & 0.87 & 0.72 & 0.67 & 0.54 & 0.47 & \textbf{0.29} & 0.47 & 0.28 & 0.60 & 0.41 \\
UOT                   & 0.68 & 0.48 & 0.52 & 0.38 & 0.46 & 0.23 & 0.41 & 0.22 & 0.52 & 0.32 \\
\textbf{\name{}} & \textbf{0.89} & \textbf{0.77} & \textbf{0.73} & \textbf{0.60} & \textbf{0.51} & \textbf{0.29} & \textbf{0.53} & \textbf{0.33} & \textbf{0.64} & \textbf{0.45} \\
\bottomrule
\vspace{-6mm}
\end{tabular*}
\end{table}

\begin{table}[htbp]
\caption{{Sim-to-real mean task scores.} P denotes the real target domain, which includes visual and physical gaps. P-OOD uses initial states outside the target-demonstration regions. Each Average is the equal-weight mean over stacking (R), mug cleanup, and ball rollover; mug cleanup includes half credit.}
\label{tab:sim2real}
\centering
\small
\setlength{\tabcolsep}{2pt}
\begin{tabular*}{\textwidth}{@{\extracolsep{\fill}}lcccccccccc@{}}
\toprule
 & \multicolumn{4}{c}{\textbf{Block Stacking}} & \multicolumn{2}{c}{\textbf{Mug Cleanup}} & \multicolumn{2}{c}{\textbf{Ball Rollover}} & \multicolumn{2}{c}{\textbf{Average}} \\
\cmidrule(lr){2-5} \cmidrule(lr){6-7} \cmidrule(lr){8-9} \cmidrule(lr){10-11}
 & \multicolumn{2}{c}{P} & \multicolumn{2}{c}{P-OOD} & & & & & & \\
\cmidrule(lr){2-3} \cmidrule(lr){4-5}
Method & R & S & R & S & P & P-OOD & P & P-OOD & P & P-OOD \\
\midrule
Source-Only              & 0.00 & 0.00 & 0.00 & 0.00 & 0.00 & 0.00 & 0.00 & 0.00 & 0.00 & 0.00 \\
Target-Only              & 0.05 & 0.00 & 0.00 & 0.00 & 0.15 & 0.00 & 0.05 & 0.00 & 0.08 & 0.00 \\
\addlinespace[2pt]
\cmidrule(lr){1-11}
\addlinespace[2pt]
Co-Training              & 0.40 & 0.25 & 0.15 & 0.15 & 0.58 & 0.18 & 0.50 & 0.40 & 0.49 & 0.24 \\
MMD                      & 0.55 & 0.40 & 0.15 & 0.05 & 0.68 & 0.20 & 0.60 & 0.30 & 0.61 & 0.22 \\
UOT                      & 0.45 & 0.30 & 0.25 & 0.20 & 0.70 & 0.33 & 0.70 & 0.50 & 0.62 & 0.36 \\
\textbf{\name{}}         & 0.75 & \textbf{0.55} & \textbf{0.45} & 0.35 & 0.78 & 0.53 & \textbf{0.85} & \textbf{0.70} & 0.79 & 0.56 \\
\textbf{MS-\name{}}      & \textbf{0.85} & \textbf{0.55} & \textbf{0.45} & \textbf{0.45} & \textbf{0.80} & \textbf{0.60} & \textbf{0.85} & \textbf{0.70} & \textbf{0.83} & \textbf{0.58} \\
\bottomrule
\vspace{-4mm}
\end{tabular*}
\end{table}

\textbf{\name{} improves sim-to-sim transfer.} Table~\ref{tab:sim2sim} shows that \name{} achieves the highest average task score of 0.64 in the \emph{texture gap} and 0.45 in the \emph{texture + viewpoint gap} over tasks that involve different motion patterns. Notably, in \emph{block stacking}, \name{} shows more robust performance when switched to the stricter S criterion; the mean relative reduction from R to S across the two visual gaps is 15.6\%, compared to 18.3\% and 28.2\% for \emph{MMD} and \emph{UOT}, respectively. This pattern is important because feature-alignment baselines already perform well; \name{} still improves over them by selecting source samples that better shape the target policy.

\textbf{Adaptive source weights improve over a tuned uniform weight.} Tuning a common source weight improves over fixed-mixture co-training, but \name{} remains stronger across the sim-to-sim tasks and gaps (Table~\ref{tab:sim2sim}). On stacking under T+V, strict success increases from 0.37 with Tuned Uniform to 0.60 with \name{}. A controlled ablation in Appendix~\ref{app:ablations} holds the $k$-NN estimator and stochastic update fixed and replaces each source discrepancy score by the mean source score. Strict success then decreases from 0.60 to 0.51 (Table~\ref{tab:beacon-ablations}). Together, these comparisons support the value of instance-level source weighting guided by local discrepancy information.

\textbf{\name{} delivers the strongest sim-to-real transfer.} Table~\ref{tab:sim2real} shows that \emph{Target-Only} policies nearly fail, and hence the limited real demonstrations are insufficient by themselves. \emph{MMD} and \emph{UOT} improve over Target-Only, but their scores drop under OOD initial states. \name{} reaches the strongest single-source performance across all three tasks, with average scores of 0.79 on P and 0.56 on P-OOD. On contact-rich ball rollover, which requires sustained gripper--ball contact and coordinated rolling and levering over a ridge, \name{} reaches 0.85 success on P and 0.70 on P-OOD, compared with 0.70 and 0.50 for UOT. Hyperparameters selected on stacking are reused for this task. These results support adaptive source weighting for transfer across combined visual and physical mismatch.

\textbf{\name{} generalizes beyond the target states seen during training.} \name{} remains robust when evaluation states move outside the region covered by the target demonstrations. Under the strict real-world stacking criterion in Table~\ref{tab:sim2real}, \name{} retains 0.35 OOD success from 0.55 in-distribution, whereas \emph{MMD} drops from 0.40 to 0.05, the lowest OOD score among the co-training methods and below plain \emph{Co-Training} (0.15). This contrast shows that an explicit feature-alignment objective alone does not ensure robustness to unseen task states.

\begin{figure}
    \centering
    \includegraphics[width=1\textwidth]{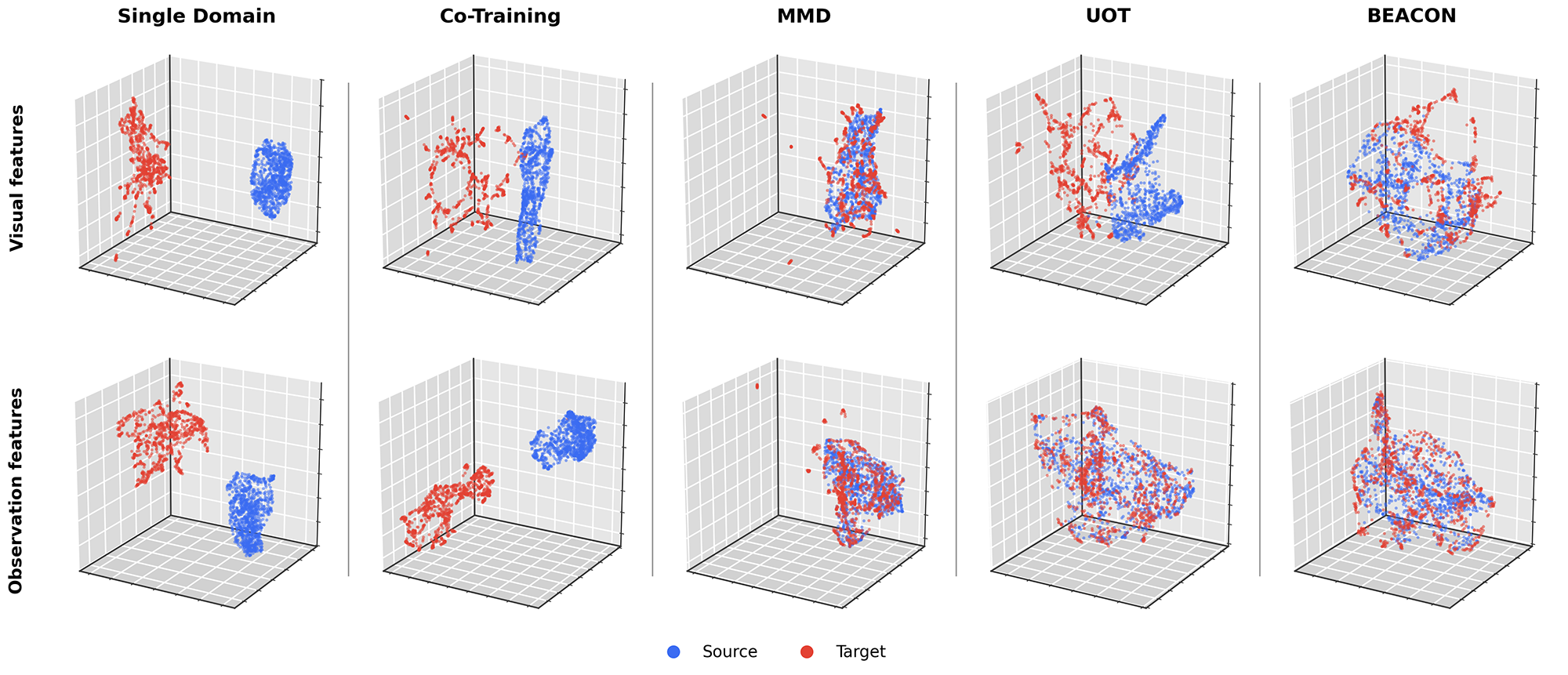}
    \vspace{-6mm}
    \caption{{UMAP visualization of latent features (\emph{block stacking}, sim-to-sim texture gap).} We visualize latent features of the image alone as well as all observations (visual + proprioception) after the encoder trunk. Source and target features come closer together in \name{} as a byproduct of discrepancy-aware co-training.}
    \label{fig:feature-alignment}
    \vspace{-6mm}
\end{figure}

\Needspace{23\baselineskip}
\begin{wrapfigure}{r}{0.45\textwidth}
    \centering
    \includegraphics[width=\linewidth]{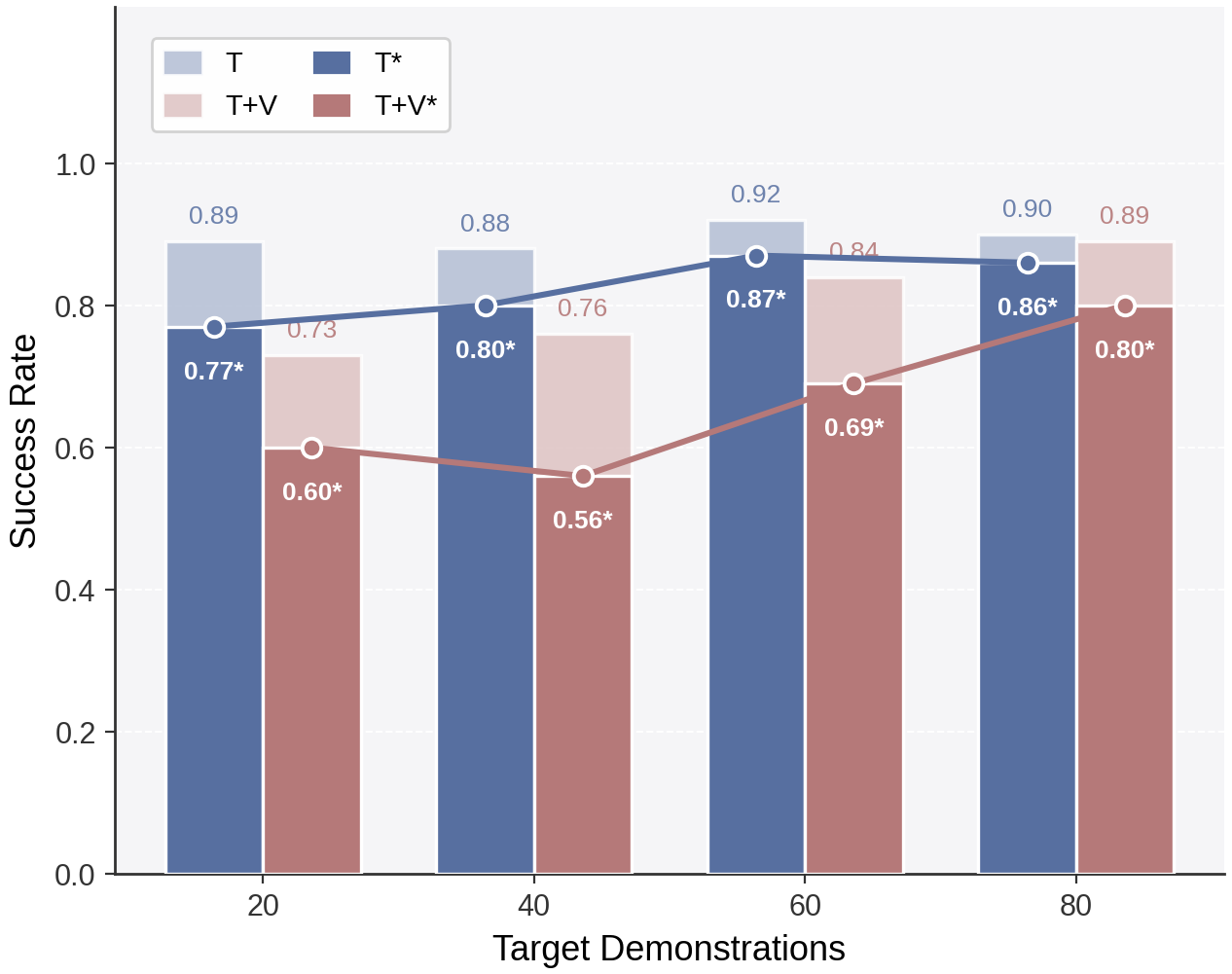}
    \caption{{Target data scaling.} The target performance shows improvement with additional target demonstrations in the sim-to-sim \emph{block stacking} setting. Starred curves show strict (S) success.}
    \label{fig:target-data-scaling}
\end{wrapfigure}

\textbf{MS-\name{} further strengthens transfer with heterogeneous sources.} With the same total source-demonstration budget, MS-\name{} raises the three-task averages to 0.83 on P and 0.58 on P-OOD (Table~\ref{tab:sim2real}). It improves relaxed stacking success and mug-cleanup scores while matching \name{} on ball rollover, indicating that heterogeneous simulation data can potentially make the learned policy both more accurate and more robust. The learned domain weights favor the source with lower measured discrepancy (Appendix~\ref{app:multi-loop}). These results support adapting how each domain and observation window contributes to policy learning.

\textbf{Closer feature alignment does not necessarily yield better transfer.} Figure~\ref{fig:feature-alignment} visualizes the source and target features learned by each method. Without an explicit feature-alignment loss, \name{} brings source and target features closer together than plain co-training, reducing RBF-MMD from 0.534 to 0.168 on stacking and from 0.460 to 0.045 on mug cleanup (Appendix~\ref{app:feature-alignment}, Table~\ref{tab:alignment-diagnostic}); this closeness arises implicitly through discrepancy-aware training. Yet the \emph{MMD} baseline is the most aligned method on both tasks, and closer alignment does not consistently translate into higher task success. Together with the OOD results, this suggests that choosing which source samples supervise the policy is more useful for transfer in our settings than minimizing feature mismatch alone.

\textbf{\name{} shows improvement as the number of target demonstrations increases.} We test the effect of scaling up the number of target demonstrations on \name{}'s performance by using 20, 40, 60, and 80 target demonstrations in the sim-to-sim \emph{block stacking} setting. Figure~\ref{fig:target-data-scaling} shows that performance generally improves with additional target demonstrations.

%% file: arxiv_source/sections/conclusion.tex

\section{Conclusion and limitations}
\label{sec:conclusion}

We presented \name{}, a theory-guided framework that recasts cross-domain co-training of generative robot policies as a discrepancy-aware importance-reweighting problem, jointly learning a diffusion policy and per-sample source weights through an objective inspired by BEA. We introduced an instance-level discrepancy proxy computed in the policy's own embedding space, stochastic alternating optimization over policy parameters and weights, and a two-level extension that combines heterogeneous source domains. Across sim-to-sim, sim-to-real, and multi-source experiments, \name{} consistently outperforms target-only, fixed-ratio co-training, and feature-alignment baselines while bringing source and target representations closer together as an implicit result of discrepancy-aware co-training. Our results suggest that choosing which source samples supervise the policy can matter more for transfer than achieving closer feature alignment. The MMD baseline attained the lowest measured RBF-MMD on both analyzed tasks but did not transfer best, and suffered a particularly sharp drop under OOD initial states in real-world stacking. This supports adaptive source weighting as a useful principle for cross-domain policy co-training.

Although we chose tasks with different motion patterns and action dynamics, our task coverage is still limited. Also, our work builds on the assumption that real data is scarce, while in reality, there might be heterogeneous but considerably larger real data available (e.g. data collected with different real-world setups, lighting conditions), which can be used as a source domain in MS-\name{}. On the theoretical side, the link between our feature-distance proxy and BEA's loss- and hypothesis-dependent labeled discrepancy remains to be established.

%% file: arxiv_source/sections/appendix/A.tex
\section{Task setups}
\label{app:tasks}
\begin{figure}
  \centering
  \includegraphics[width=\textwidth]{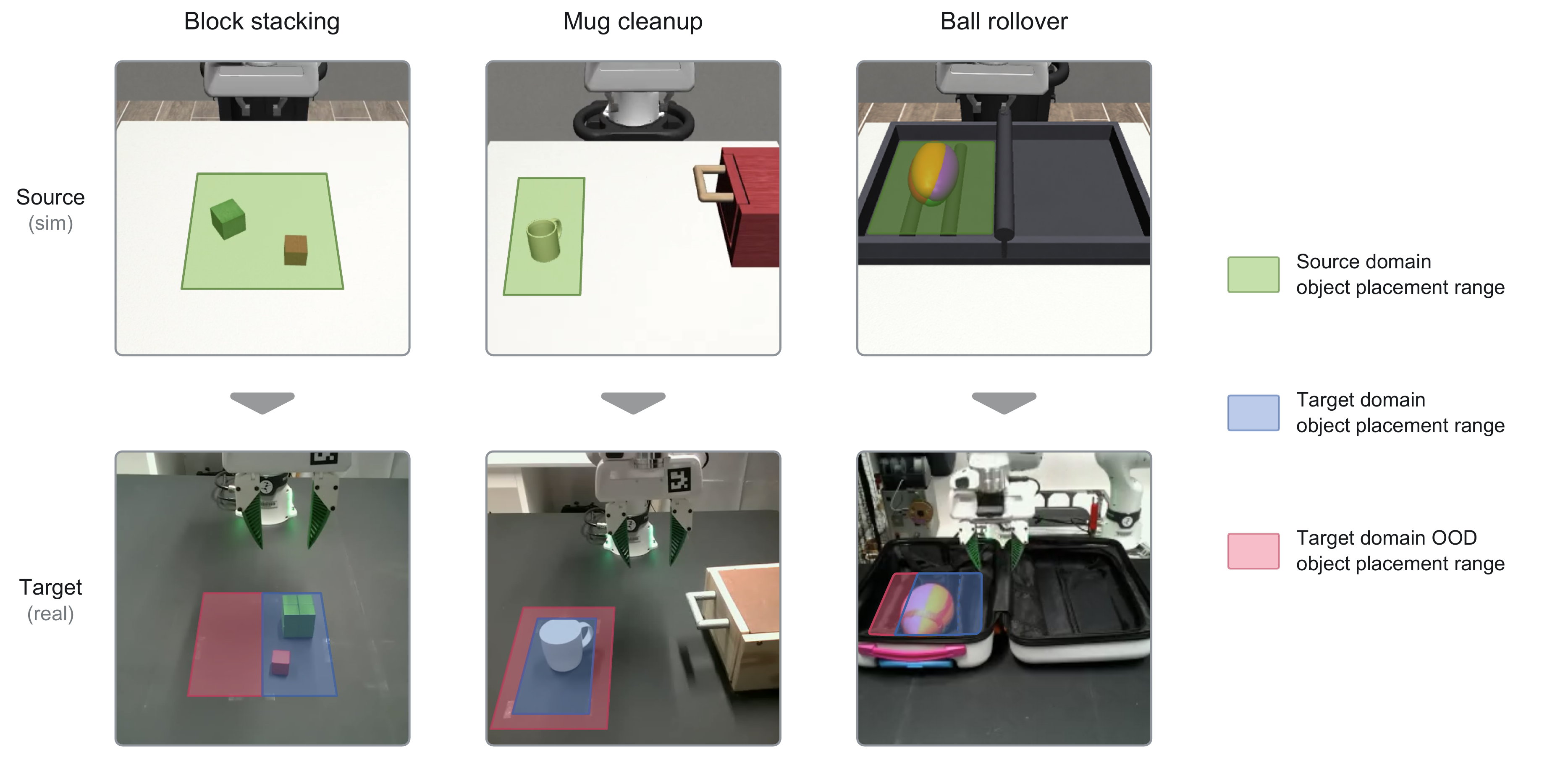}
  \caption{Object placement range for OOD evaluation. The green region shows the placement range for source demonstrations; the blue region shows the placement range for target demonstrations, and the red regions are the OOD evaluation regions. Each column shows one task, with simulation above real-world experiments. For ball rollover, placement regions are restricted to the left luggage compartment: simulation covers its full interior, while real demonstrations cover only the middle and right sections; the uncollected left section is held out as OOD and requires an extra left-to-middle motion before moving to the right section.}
  \label{ood}
  \vspace{-3mm}
\end{figure}

We conduct comprehensive sim-to-sim, sim-to-real, and multi-source experiments on tasks that involve different motion patterns.

\textbf{Evaluation and scoring.} Each simulated policy is evaluated on 100 rollouts and each real-robot policy on 20 rollouts. Initial states are sampled independently across rollouts and matched across methods for each task. Each simulated rollout uses a distinct evaluation seed, separate from the training seed. Mug cleanup awards 1 for complete success and 0.5 for partial completion; the reported values are therefore mean task scores. Stacking, threading, and ball rollover use binary success. The task objectives and success criteria are defined below.

\textbf{Block stacking:} grasp the smaller cube (A) and stack it on the larger one (B). This task involves simple pick-and-place motions. Both cubes' $(x,y)$ positions are independently randomized over a square region; OOD evaluation uses a disjoint region shown in Figure~\ref{ood}. Relaxed (R) success requires cube A to be stacked on cube B at any frame; strict (S) success requires the stack to persist for at least five consecutive frames or occur at the final frame.

\textbf{Mug cleanup:} drag and open the drawer, pick up a mug and place it upright inside the drawer, and then close the drawer. This task combines pulling and pushing the drawer with picking and placing the mug. The mug's $(x,y)$ positions are randomized over a rectangular region; OOD evaluation uses a spatially disjoint region. The task is considered successful when the mug is inside the drawer, upright, and the drawer is closed. From a full score of 1, a mug that is not upright and a drawer that is not fully closed each incur a 0.5-point deduction. A failed attempt scores zero.

\textbf{Threading:} grasp a thin rod/thread and insert it through a small receptacle. This task requires precise insertion with sub-centimeter alignment. The rod's $(x,y)$ and yaw are randomized over a rectangular region. The task is considered successful when the rod/thread is inserted through the receptacle.

\textbf{Ball rollover:} a suitcase lies flat on the table, with a football at a randomized position on one side. The gripper rolls the ball to the central ridge, reaches its base, and levers it into the opposite compartment. This contact-rich task requires sustained gripper--ball contact through the rolling and levering sequence, with the ridge nearly as tall as the ball. Success requires moving the ball across the ridge into the opposite compartment. The multi-source experiment uses the same two simulation domains and total source-demonstration budget as the other sim-to-real tasks.

%% file: arxiv_source/sections/appendix/B.tex
\section{Algorithm and training details}
\label{app:algorithm}

This section describes how \name{} is implemented and trained in practice. We start
with the single-source training loop (Section~\ref{app:single-loop}), then detail the
$k$-nearest-neighbor discrepancy estimator that supplies the per-sample $d_i$ values
(Section~\ref{app:knn}), and finally describe the multi-source extension MS-\name{}
(Section~\ref{app:multi-loop}). Implementation-level choices and default hyperparameters appear in Section~\ref{app:training-details}.

\subsection{Single-source training loop}
\label{app:single-loop}

The objective in \eqref{eq:beacon-objective} is jointly nonconvex in
$(\theta,\mathbf{q})$: it is convex in $\mathbf{q}$ for fixed $\theta$ but nonconvex in
$\theta$ for fixed $\mathbf{q}$ because $\ell_{\mathrm{BC}}$ is the diffusion
behavior-cloning loss of a deep policy. We therefore optimize it by alternating
a weight-refresh phase and a policy-training phase (Algorithm~\ref{alg:beacon}). On refresh epochs, the $\mathbf{q}$-phase fixes $\theta$
and takes one sweep of projected subgradient steps over $\mathbf{q}$, followed by a
single global projection that enforces the box and sum constraints on the weights.
In every epoch, the $h$-phase fixes $\mathbf{q}$ and takes a configurable number of minibatch
steps on the $\mathbf{q}$-weighted policy loss. Decoupled weight decay is rescaled to
$\gamma\,\|\mathbf{q}\|_\infty$ during the $h$-phase, which serves as an optimizer-level analogue of the
capacity term $\gamma\,\|\mathbf{q}\|_\infty\,R(\theta)$ in
\eqref{eq:beacon-objective}, without an
explicit gradient through $R(\theta)$.

\begin{algorithm}[ht]
\caption{\name{}: single-source training loop}
\label{alg:beacon}
\begin{algorithmic}[1]
\Require Combined dataset $\mathcal{D}$ with source/target indices $S,T$; policy $\pi_\theta$; reference $\mathbf{r}$; coefficients $(\lambda_d,\lambda_1,\lambda_2,\gamma,\alpha)$; step sizes $(\eta_q,\eta_\theta)$; bounds $[\underline q_i, q_{\max}]$; $q$-refresh period $K_q$; $h$-steps per epoch $S_h$; epochs $E$.
\State Initialize $\theta$ and $q_i\gets r_i$ for all $i\in S\cup T$.
\For{$e = 1,\dots,E$}
  \If{$e=1$ \textbf{or} $e \bmod K_q = 0$}
    \State Refresh $\{d_i\}_{i\in S}$ via Algorithm~\ref{alg:knn}.\label{line:refresh}
    \For{each minibatch $\mathcal{B}\subset\mathcal{D}$ in a sequential sweep}
      \State Evaluate $\{\ell_i(\theta)\}_{i\in\mathcal{B}}$ (no backprop).
      \For{$i\in\mathcal{B}$}
        \State $g_i \gets \ell_i + \lambda_d d_i + \gamma R(\theta)\,v_i + \lambda_1\,\mathrm{sign}(q_i - r_i) + 2\lambda_2 q_i$ \Comment{cf.~\eqref{eq:q-grad}}
        \State $q_i \gets \mathrm{clip}_{[\underline q_i,\,q_{\max}]}\!\big(q_i - \eta_q g_i\big)$
      \EndFor
    \EndFor
    \State Project $\mathbf{q}$ onto $\{q_i\!\in\![\underline q_i,q_{\max}],\;\sum_i q_i = n + \alpha m\}$ by bisection.\label{line:bisect}
  \EndIf
  \Statex \hspace{0.3em}
  \For{$s = 1,\dots,S_h$}
    \State Sample minibatch $\mathcal{B}$ from the source--target sampling mixture.
    \State $\theta \gets \mathrm{AdamW}\!\big(\theta,\,\nabla_\theta\,\tfrac{1}{|\mathcal{B}|}\!\sum_{i\in\mathcal{B}} q_i\,\ell_i(\theta);\;\eta_\theta\big)$.
  \EndFor
\EndFor
\State \Return $\theta$.
\end{algorithmic}
\end{algorithm}

\subsection{Per-sample \texorpdfstring{$k$}{k}-NN discrepancy estimator}
\label{app:knn}

We instantiate the source--target discrepancy proxy $d_i$ used in
\eqref{eq:beacon-objective} as a per-sample $k$-nearest-neighbor distance in the
embedding space of the policy's observation encoder $\phi_\theta$, with self-normalization
by the within-target spread (Algorithm~\ref{alg:knn}). The embedding concatenates visual and proprioceptive features over the observation window. The normalization makes the
scale of $d_i$ approximately invariant to a common rescaling of the embedding
space. Because $\phi_\theta$ co-evolves with the policy, we recompute $\{d_i\}_{i\in S}$ at the
beginning of every $\mathbf{q}$-update phase (line~\ref{line:refresh} of
Algorithm~\ref{alg:beacon}); when the target sample is small ($n<k$), the effective
neighborhood size collapses to $k_{\mathrm{eff}}\!=\!\min(k,n)$.

\begin{algorithm}[ht]
\caption{Per-sample $k$-NN discrepancy in policy embedding space}
\label{alg:knn}
\begin{algorithmic}[1]
\Require Inputs $\{x_i\}_{i\in S\cup T}$; encoder $\phi_\theta$; neighbors $k$ (with $k_{\mathrm{eff}}\!\!=\!\!\min(k,n)$).
\State $z_i \gets \phi_\theta(x_i)$ for all $i\in S\cup T$ \Comment{single forward pass through $\phi_\theta$}
\State $Z \gets$ mean $k_{\mathrm{eff}}$-NN distance among target embeddings $\{z_j\}_{j\in T}$, excluding self-matches.
\For{$i\in S$}
  \State $d_i \gets \big(\text{mean distance from $z_i$ to its $k_{\mathrm{eff}}$ nearest target embeddings}\big)\,/\,Z$.
\EndFor
\State \Return $\{d_i\}_{i\in S}$.
\end{algorithmic}
\end{algorithm}

\subsection{Multi-source training loop}
\label{app:multi-loop}

For multi-source training we factor the per-sample weight as
$q_i\!=\!w_{s(i)}\widetilde q_i$ for $i\in S$ and $q_j\!=\!\widetilde q_j$ for
$j\in T$, where $\mathbf{w}\in\Delta^{K-1}$ is the simplex-constrained vector of
source-domain weights and $\widetilde{\mathbf{q}}$ is the nonnegative within-source/target
weight vector. The multi-source objective augments \eqref{eq:beacon-objective} with
an additional $L_1/L_2$ regularizer on $\mathbf{w}$ around a reference $\mathbf{w}_0$:
\begin{equation}
\min_{\theta,\,(\mathbf{w},\widetilde{\mathbf{q}})\in\mathcal{M}_b}\;
\mathcal{F}_{\mathrm{single}} \big(\theta,\,\mathbf{q}(\mathbf{w},\widetilde{\mathbf{q}})\big)
\;+\; \rho_1\|\mathbf{w}-\mathbf{w}_0\|_1 \;+\; \rho_2\|\mathbf{w}\|_2^2,
\label{eq:msbeacon-objective}
\end{equation}
where $\mathcal{F}_{\mathrm{single}}$ is the single-source \name{} objective
\eqref{eq:beacon-objective} evaluated on the composite weights
$\mathbf{q}(\mathbf{w},\widetilde{\mathbf{q}})$. The feasible set $\mathcal{M}_b$ requires $\mathbf{w}\in\Delta^{K-1}$, $0\le\widetilde q_i\le q_{\max}$ for $i\in S$, $q_T^{\min}\le\widetilde q_j\le q_{\max}$ for $j\in T$, and $\sum_{i\in S}w_{s(i)}\widetilde q_i+\sum_{j\in T}\widetilde q_j=b$. This induces three blocks of variables (Algorithm~\ref{alg:msbeacon}). On refresh epochs, we perform a
\emph{$w$-step} that uses a single full-dataset gradient, scaled by $1/\bar m$ with $\bar m=m/K$, followed by clipping to $[0,1]$ and
projection onto $\Delta^{K-1}$; a \emph{$\widetilde q$-step} that takes one minibatch
sweep of projected subgradient on the within-source weights and a separate update on
the target weights; and a final \emph{joint projection} of the composite vector
$\mathbf{q}$ onto the box-and-sum constraint set with source bounds $0\le q_i\le w_{s(i)}q_{\max}$ and target bounds $q_T^{\min}\le q_j\le q_{\max}$, which is then split back into
$\widetilde{\mathbf{q}}$ via the change of variables above. The $h$-step runs in every epoch and is identical
to the single-source case but uses the composite weights $q_i\!=\!w_{s(i)}\widetilde q_i$
when forming the weighted minibatch loss. The $h$-step uses the same optimizer-level weight decay as Algorithm~\ref{alg:beacon}.
The multi-source $w$- and $\widetilde q$-updates omit the capacity subgradient and use
their respective regularizers $(\rho_1,\rho_2)$ and $(\lambda_1,\lambda_2)$. Because the two sources in our experiments contain nearly equal numbers of observation windows (within $0.31\%$), the common scale matches normalizing each domain's term by its own window count to within $0.31\%$, so the learned domain weights are not driven by source size.

\begin{algorithm}[ht]
\caption{MS-\name{}: multi-source training loop}
\label{alg:msbeacon}
\begin{algorithmic}[1]
\Require $K$ source datasets $\{\mathcal{D}_{\mathrm{src},k}\}$ with assignment $s:S\!\to\!\{1,\dots,K\}$; target $\mathcal{D}_\mathrm{tgt}$; reference $\mathbf{w}_0\!=\!(1/K)\mathbf{1}$; domain regularizers $(\rho_1,\rho_2)$; step sizes $(\eta_w,\eta_T)$; remaining hyperparameters as in Algorithm~\ref{alg:beacon}.
\State Initialize $\theta$, $w_k\!\gets\!1/K$, and $\widetilde{\mathbf{q}}$ from $\mathbf{r}$; let $q_i\!=\!w_{s(i)}\widetilde q_i$ for $i\in S$ and $q_j\!=\!\widetilde q_j$ for $j\in T$.
\For{$e = 1,\dots,E$}
  \If{$e=1$ \textbf{or} $e\bmod K_q = 0$}
    \State Refresh $\{d_i\}_{i\in S}$ via Algorithm~\ref{alg:knn}; collect $\{\ell_i(\theta)\}_{i\in S\cup T}$ in one sequential sweep of $\mathcal{D}$.
    \State \emph{$w$-step:} $\mathbf{w} \gets \Pi_{\Delta^{K-1}}\!\big(\mathrm{clip}_{[0,1]}(\mathbf{w} - (\eta_w/\bar m)\,\nabla_{\mathbf{w}}\mathcal{F})\big)$, where $\nabla_{\mathbf{w}}\mathcal{F}$ is the closed-form gradient of \eqref{eq:msbeacon-objective} in $\mathbf{w}$ (with the $\arg\max$ capacity term excluded).
    \State \emph{$\widetilde q$-step:} for each minibatch $\mathcal{B}$ in a second sequential sweep of $\mathcal{D}$, take a projected-subgradient step on $\widetilde q_i$ ($i\in\mathcal{B}\cap S$, step size $\eta_q$) and $\widetilde q_j$ ($j\in\mathcal{B}\cap T$, step size $\eta_T$) using the per-sample subgradients of \eqref{eq:msbeacon-objective} under the change of variables $q=w\widetilde q$.
    \State \emph{Joint projection:} form $\mathbf{q}$ and project onto the box-and-sum set using source upper bounds $w_{s(i)}q_{\max}$ and target bounds $[q_T^{\min},q_{\max}]$, then read back $\widetilde q_i\!=\!q_i/w_{s(i)}$ for $i\in S$ and $\widetilde q_j\!=\!q_j$ for $j\in T$.
  \EndIf
  \State \emph{$h$-step:} run the $h$-update of Algorithm~\ref{alg:beacon} with composite weights $q_i\!=\!w_{s(i)}\widetilde q_i$ ($i\in S$) and $q_j\!=\!\widetilde q_j$ ($j\in T$).
\EndFor
\State \Return $\theta$.
\end{algorithmic}
\end{algorithm}

\textbf{Domain-weight diagnostics.}
In the stacking and mug-cleanup multi-source runs, each source contributes 250 demonstrations and receives equal sampling mass, so the learned domain weights are not explained by source size or sampling. On block stacking, the weights select the source with the lower mean discrepancy almost exclusively ($0.98$ versus $0.02$). On mug cleanup, the two sources are nearly equidistant from the target (mean discrepancies $4.276$ and $4.281$), and the preference is correspondingly soft ($0.70$ versus $0.30$, again favoring the closer source). With two sources, solving the domain-weight subproblem exactly reproduces the learned selections, and none of 129 tested settings of $(\rho_1,\rho_2,\lambda_1,\lambda_2,\lambda_d)$ reverses them. Within each source, the learned weights $\widetilde q_i$ correlate strongly with the discrepancy proxy ($-0.91$ to $-0.99$) but only weakly with the imitation loss ($-0.04$ to $-0.29$), suggesting that within-domain weighting mainly suppresses mismatched windows rather than difficult ones.

\subsection{Implementation notes and essential hyperparameters}
\label{app:training-details}

The policy backbone, optimizer, and learning-rate schedule follow the public
Diffusion Policy implementation~\citep{chi2023diffusion} and are kept identical
across all baselines and \name{} variants. Training windows use stride one within each demonstration, repeating endpoints for padding, with observation, prediction, and execution horizons of 2, 16, and 8. Policy minibatches are drawn from a fixed source--target sampling mixture, uniformly within each domain, and the learned weights multiply the sampled losses. The \name{}-specific knobs that matter most in practice are the loss-coefficient triple $(\lambda_1,\lambda_2,\lambda_d)\!\approx\!(10^{-2},10^{-2},10^{-1})$, the per-sample upper bound $q_{\max}\!\approx\!5$ together with the target floor
$q_T^{\min}\!\in\!\{0.05,0.10\}$ that prevents target weights from collapsing, the
sum budget $\alpha\!\in\![0.4,0.5]$ in $\sum_i q_i = n + \alpha m$, the $k$-NN
neighborhood $k\!\in\!\{4,5,6\}$, and the $q$-step size $\eta_q\!=\!10^{-2}$. We use
$K_q\!=\!1$ for sim-to-sim and $K_q\!=\!2$ for sim-to-real, with an initial refresh before policy training. For MS-\name{} the
additional knobs are the source-level regularizers
$(\rho_1,\rho_2)\!\approx\!(10^{-2},10^{-2})$ and the $w$-step size
$\eta_w\!\in\![10^{-2},10^{-1}]$. Sensitivity around these defaults is reported in
Section~\ref{app:hyperparam}.

%% file: arxiv_source/sections/appendix/C.tex
\section{Alternative \texorpdfstring{$\mathbf{q}$}{q}-update and discrepancy estimators}
\label{app:ablations}

The default \name{} loop in Algorithm~\ref{alg:beacon} couples a minibatch projected
subgradient on $\mathbf{q}$ with a $k$-NN discrepancy estimator. We retain one
alternative for $\mathbf{q}$ and two alternatives for the discrepancy term, all of which we
ablate against the defaults.

\textbf{Full-convex inner solve for $\mathbf{q}$.}
A solver closer to~\citet{awasthi2023best}'s alternating minimization solves the
$\mathbf{q}$ subproblem in closed form for fixed $\theta$. Treating $t\!=\!\|\mathbf{q}\|_\infty$ as
auxiliary, the inner problem at fixed $(\theta, t)$ becomes separable across
coordinates and admits the closed-form solution
\begin{equation}
q_i^\star(t) \;=\; \mathrm{clip}_{[\underline q_i,t]}\!\Big(r_i + \mathcal{S}_{\lambda_1/(2\lambda_2)}\!\big(z_i(\theta) - r_i\big)\Big),\qquad z_i(\theta) := \frac{\mu-\ell_i(\theta) - \lambda_d\, d_i}{2\lambda_2},
\label{eq:qfull}
\end{equation}
where $\mathcal{S}_\tau(u) = \mathrm{sign}(u)\max(|u|-\tau, 0)$ is the soft-threshold operator and $\mu$ is the Lagrange multiplier of the sum constraint. The convex outer problem $\min_{t\ge 0}\,[\,\mathrm{inner}(t) + \gamma\, R(\theta)\, t\,]$ is one-dimensional and convex, and we solve it by ternary search to floating-point precision. The per-sample sum constraint $\sum_i q_i = n + \alpha m$ is incorporated via a second bisection on $\mu$, exactly as in line~\ref{line:bisect} of Algorithm~\ref{alg:beacon}. In this ablation, the costs $\ell_i(\theta)+\lambda_d d_i$ are divided by their maximum whenever it exceeds one.

\textbf{Localized labeled discrepancy.}
The first BEA-style alternative is the most faithful to the original derivation of $\mathrm{dis}(P,Q)$ in~\citet{awasthi2023best}, with the local-class refinement of~\citet{cortes2019adaptation,zhang2020localized}. As a one-time pre-processing step before the main \name{} loop, we train a target-only reference policy $\pi_{\theta_\mathrm{ref}}$ on $\mathcal{D}_\mathrm{tgt}$, partition its parameters into the (frozen) observation encoder $\theta_\mathrm{enc}$ and the (free) noise-prediction parameters $\theta_\mathrm{free}$, and define a parameter-space ball over the free part,
\begin{equation*}
\mathcal{H}_\mathrm{loc} \;=\; \bigl\{\pi_{\theta'} : \theta'_\mathrm{enc}=\theta_\mathrm{enc,ref},\;\;\|\theta'_\mathrm{free} - \theta_\mathrm{free,ref}\|_2 \le \rho\bigr\},
\end{equation*}
with radius $\rho=0.05\|\theta_{\mathrm{free,ref}}\|_2$. We then run projected gradient ascent on $\theta'\in\mathcal{H}_\mathrm{loc}$ to maximize the empirical labeled discrepancy
\begin{equation}
\widehat{\mathrm{dis}}(P,Q;\theta') \;=\; \frac{1}{n}\sum_{j\in T}\ell_\mathrm{BC}(\theta';x_j,y_j) \;-\; \frac{1}{m}\sum_{i\in S}\ell_\mathrm{BC}(\theta';x_i,y_i),
\label{eq:dhat_obj}
\end{equation}
augmented with a soft proximity penalty $-\beta\,\|\theta'_\mathrm{free}-\theta_\mathrm{free,ref}\|_2^2$ ($\beta\!>\!0$) that prevents gradient steps from concentrating mass on the boundary of the ball; the hard ball constraint is enforced by Euclidean projection of $\theta'_\mathrm{free}$ after each step. Every few steps we evaluate $\widehat{\mathrm{dis}}(P,Q;\theta')$ on the full labeled data, averaging over a small number of independent diffusion-noise/timestep draws to reduce the upward bias of taking a maximum over noisy estimates, and track the running maximum. We set $\hat d$ to this maximum, fix $d_i = \hat d\,\mathbf{1}[i\in S]$ and $\lambda_d = 1$, and freeze them for the entire \name{} run, so line~\ref{line:refresh} of Algorithm~\ref{alg:beacon} becomes a no-op.

\textbf{Domain-classifier discrepancy.}
The second alternative replaces the domain-level loss difference~\eqref{eq:dhat_obj} with a per-source instance-level mismatch score derived from a binary domain classifier on the policy's observation-encoder features, an instance-level analogue of the $d_\mathcal{A}$-style discriminator estimators of~\citet{ben2010theory,ganin2016domain}. At each refresh we (i)~extract the current embeddings $\{\phi_\theta(x_i)\}_{i\in S\cup T}$ in a single forward pass through $\phi_\theta$, (ii)~draw a class-balanced training set by random source undersampling so that source and target contribute equally (since typically $m\!\gg\!n$), (iii)~fit a logistic-regression discriminator $g_\psi : \phi_\theta(x)\!\mapsto\![0,1]$ with target labeled $1$ and source labeled $0$, and (iv)~apply $g_\psi$ to \emph{all} $m$ source embeddings (not just the balanced subset) to obtain
\begin{equation}
d_i \;=\; 1 - g_\psi\!\big(\phi_\theta(x_i)\big) \;=\; \widehat{\Pr}\!\big(\mathrm{source}\mid x_i\big),\qquad i\in S,\qquad d_j=0 \text{ for } j\in T,
\label{eq:dhat_classifier}
\end{equation}
so that source samples that look target-like ($g_\psi\!\to\!1$) get $d_i\!\to\!0$ and clearly-source samples ($g_\psi\!\to\!0$) get $d_i\!\to\!1$. We additionally log $\mathrm{AUC}(g_\psi)$ on the full pooled $S\cup T$ as a global indicator of domain separability, but use only the per-instance $d_i$ from~\eqref{eq:dhat_classifier} inside the \name{} objective. Unlike the localized estimator, both $g_\psi$ and $\{d_i\}_{i\in S}$ here are recomputed in place of line~\ref{line:refresh} of Algorithm~\ref{alg:beacon} so that the score tracks the encoder $\phi_\theta$ as it evolves.

\begin{table}[t]
    \caption{Average target success rate of eight \name{} variants on sim-to-sim \emph{block stacking} with \emph{texture + viewpoint gap}, evaluated on the strict S criterion. The two $k$-NN columns hold the estimator fixed while varying discrepancy granularity. The bolded entry is the default configuration used in Section~\ref{sec:experiments}.}
    \label{tab:beacon-ablations}
    \centering
    \small
    \begin{tabular*}{\textwidth}{@{\extracolsep{\fill}}lcccc@{}}
    \toprule
                                    & \textbf{Localized} & \multicolumn{2}{c}{\textbf{$k$-NN}} & \textbf{Classifier} \\
    \cmidrule(lr){3-4}
    \textbf{$\mathbf{q}$-update}    & \textbf{labeled} & \textbf{Source mean} & \textbf{Instance-level} & \\
    \midrule
    Full-convex inner solver        & 0.58 & 0.48 & 0.52          & 0.54 \\
    Stochastic update (default)     & 0.58 & 0.51 & \textbf{0.60} & 0.58 \\
    \bottomrule
    \end{tabular*}
    \end{table}

\textbf{Empirical comparison.}
Table~\ref{tab:beacon-ablations} compares eight variants across two $\mathbf{q}$-update rules and three estimator families, with two score granularities for $k$-NN. To isolate discrepancy granularity, the source-mean variant replaces each source score $d_i$ by $\bar d=\frac{1}{m}\sum_{i\in S}d_i$, while retaining the same estimator and weight-update rule. The sample weights remain adaptive; only the discrepancy scores are shared across source samples. With the stochastic update, instance-level scores achieve $0.60$ strict stacking success, compared with $0.51$ for source-mean scores; with the full-convex update, the corresponding scores are $0.52$ and $0.48$. The localized and classifier columns separately compare estimator families. The default stochastic update with instance-level $k$-NN scores achieves the highest score.

%% file: arxiv_source/sections/appendix/D.tex
\section{Hyperparameter sensitivity analysis}
\label{app:hyperparam}

We conduct a sensitivity analysis over the training hyperparameters, including the $L_1$ deviation coefficient $\lambda_1$, the $L_2$ shrinkage $\lambda_2$, the discrepancy weight $\lambda_d$, and the sum budget $\alpha$. We sweep one hyperparameter at a time, retaining the value with the highest target success rate before moving to the next. A single sweep on sim-to-sim \emph{block stacking} is reused unchanged for \emph{mug cleanup} and \emph{threading}; a separate sim-to-real stacking sweep is reused for \emph{mug cleanup} and \emph{ball rollover}. OOD evaluations are never used for selection. The coefficient $\lambda_2$ discourages concentrated weights, $\lambda_d$ scales the discrepancy penalty, and $\alpha$ sets the total-weight budget through the projection in line~\ref{line:bisect} of Algorithm~\ref{alg:beacon}. Under the stated target floors and fixed total-weight budget, the inherited $L_1$ term is constant over the feasible set.

The sim-to-sim sweep identifies a stable region with $\lambda_1\!=\!0.01$, $\lambda_2\!\in\!\{0.01,0.001\}$, $\lambda_d\!\in\!\{0.1,0.05\}$, and $\alpha\!\in\![0.3,0.5]$. Sim-to-real uses smaller $\lambda_d$ to accommodate larger discrepancy scores and smaller $\alpha$ to reduce the total-weight budget. Stored-checkpoint audits of the default runs found no oscillation or zeroed source weights.

For a new transfer setting, the self-normalized discrepancy scores make the two active knobs readable from the data. Since $d_i\approx 1$ means that a source window lies within the typical target spread, $\alpha$ can be set from the empirical histogram of $d_i$ (roughly the fraction of source windows with $d_i$ near or below $1$), and $\lambda_d$ at the scale that makes $\lambda_d d_i$ comparable to a typical target behavior-cloning loss; both quantities follow from one forward pass of a target-only policy. Where target data can be spared, a held-out target split selected on validation behavior-cloning loss serves the same purpose. We offer this as a heuristic starting point; the reported runs used the sweep protocol above.

For the Tuned Uniform baseline, we sweep the common source weight over $\{0.1,0.5,0.9,2,8,16\}$ on block stacking, select 8, and reuse it across all sim-to-sim tasks and visual gaps.

%% file: arxiv_source/sections/appendix/E.tex
\section{Feature alignment}
\label{app:feature-alignment}

\begin{figure}
  \centering
  \includegraphics[width=1\textwidth]{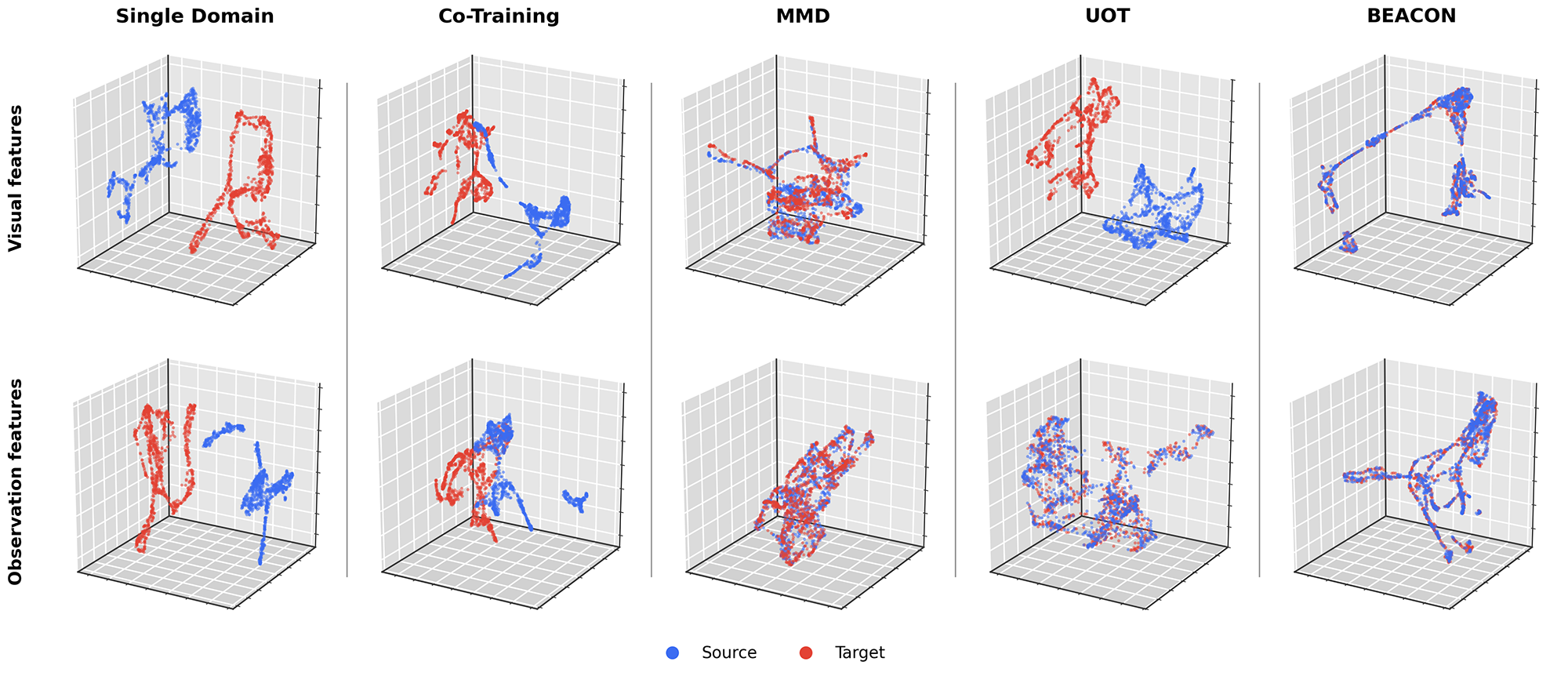}
  \caption{{UMAP visualization of latent features (\emph{mug cleanup}, sim-to-sim texture gap).} As in Figure~\ref{fig:feature-alignment}, the top row shows features of the image alone and the bottom row all observations (visual + proprioception) after the encoder trunk.}
  \label{fig:feature-alignment-mug}
\end{figure}

Figure~\ref{fig:feature-alignment-mug} complements the stacking visualization with latent features from \emph{mug cleanup}. We quantify source--target alignment on the frozen policy encoders from the sim-to-sim texture-gap experiments, using the full observation features (visual and proprioceptive), with the RBF-MMD between source and target features and the held-out error of a linear domain classifier (Table~\ref{tab:alignment-diagnostic}). Without an explicit alignment loss, \name{} reduces RBF-MMD relative to co-training by about $3\times$ on block stacking and $10\times$ on mug cleanup, and its features are harder to separate by domain (probe error $0.13$ and $0.41$, versus $0.00$ for co-training). The MMD baseline, which minimizes this discrepancy directly, attains the lowest values on both tasks without the highest task success. Holding \name{}'s trained encoder fixed, reweighting the source features by $\mathbf q$ changes the RBF-MMD by less than $0.01$ on both tasks (paired bootstrap intervals include zero), so the closer alignment resides in the trained encoder and does not depend on reweighting at evaluation time.

\begin{table}[ht]
\centering
\caption{Source--target feature alignment on frozen policy encoders (sim-to-sim texture gap, full observation features, sampled source windows unweighted). RBF-MMD: lower values indicate closer alignment. $\epsilon$: held-out error of a linear domain classifier ($0.5$ is chance); $\hat A=\max\{0,2(1-2\epsilon)\}$ is the proxy $\mathcal A$-distance, with lower values indicating less separable features.}
\label{tab:alignment-diagnostic}
\small
\begin{tabular}{lcccc}
\toprule
 & Co-Training & MMD & UOT & \name{} \\
\midrule
RBF-MMD, block stacking & 0.534 & \textbf{0.082} & 0.124 & 0.168 \\
RBF-MMD, mug cleanup & 0.460 & \textbf{0.028} & 0.041 & 0.045 \\
$\epsilon$ / $\hat A$, block stacking & 0.00 / 2.00 & 0.58 / 0.00 & 0.00 / 1.99 & 0.13 / 1.49 \\
$\epsilon$ / $\hat A$, mug cleanup & 0.00 / 2.00 & 0.52 / 0.00 & 0.00 / 2.00 & 0.41 / 0.37 \\
\bottomrule
\end{tabular}
\end{table}

%% file: arxiv_source/sections/appendix/F.tex
\section{Source weight evolution during training}
\label{app:weight_evolution}

\begin{figure}
  \centering
  \includegraphics[width=1\textwidth]{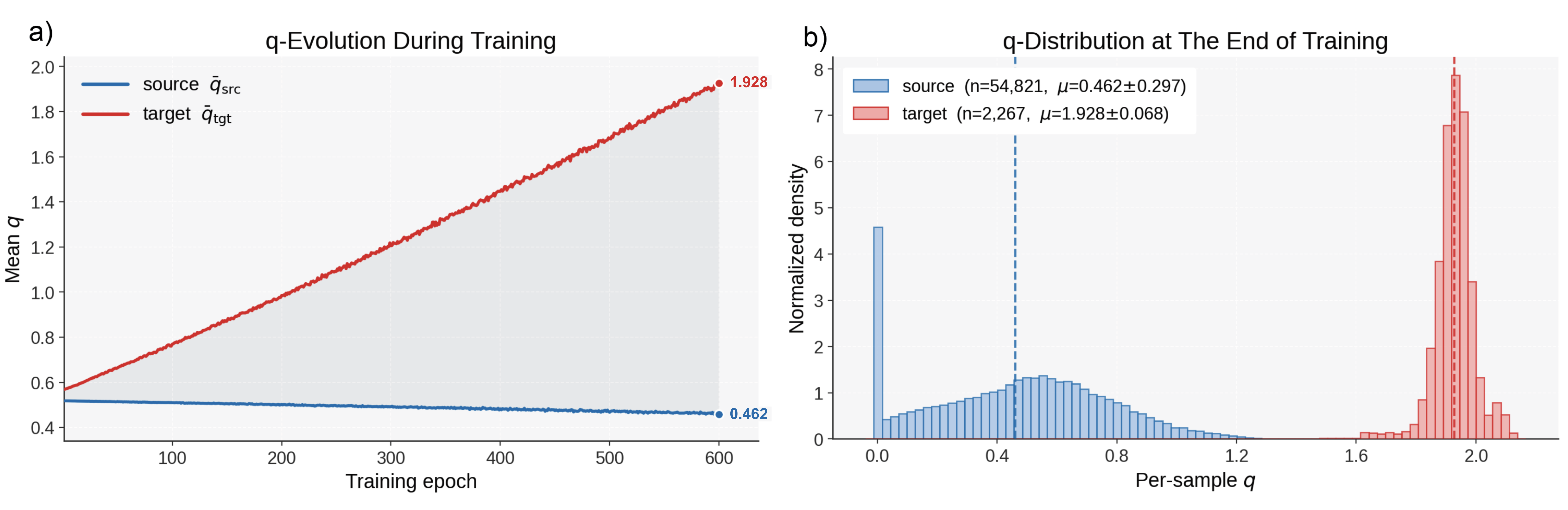}
  \caption{a) $\mathbf{q}$-evolution during training. The average $\mathbf{q}$ values for source windows (blue) decrease over time, while the average $\mathbf{q}$ values for target windows (red) increase at a steeper rate. b) $\mathbf{q}$-distribution at the end of training. Density is normalized to the total number of samples in each domain so the two very different sample counts are comparable. Both plots are for the \emph{block stacking} task.}
  \label{q-analysis}
\end{figure}

The $\mathbf{q}$-evolution during training and weight distributions for \emph{block stacking} task are shown in Figure~\ref{q-analysis}. The target mean weight rises steeply during training from $\sim\!0.5$ to $\sim\!1.9$ over $600$ epochs, while the source mean weight drifts from $0.51$ to $0.46$. At the end of training, target weights are tightly concentrated (mean $1.93$, std $0.07$), well below the cap $q_{\max}=5$. Their increase reflects redistribution under the total-weight budget. Across all 12 reported runs, no weight reaches either edge of its box, and the largest weight is 11--63\% of $q_{\max}$. After normalization by $\sum_i q_i$, all weights lie in $[0,1]$; because the admitted source mass exceeds the target mass, however, the normalized weights do not satisfy BEA's condition $\|\mathbf q-\mathbf p_0\|_1<1$ (Theorem~\ref{thm:discrepancy-generalization}). Source weights are more dispersed (mean $0.46$, std $0.30$), with a near-zero peak and a broader mode around $0.5$.

Additional audits show that weighting acts mainly within demonstrations, with 78--96\% of the source-weight variance lying within individual trajectories. Stretches with $q_i<0.05$ have a median length of five to six frames. In sim-to-real stacking, high-weight windows tend to occur during slow placement near the end of the episode, while low-weight windows occur during faster transport. This pattern shows how the learned weights distinguish task phases within a demonstration.

%% file: arxiv_source/sections/appendix/G.tex
\section{Compute resources}
\label{app:compute}

Experiments use a single NVIDIA L40S GPU with 48\,GB of memory per run, and a typical \name{} run completes within 24 hours. In the reported block-stacking profile, a \name{} epoch takes 40.3\,s, compared with 38.7\,s for co-training. Across profiled settings, the per-epoch ratio ranges from $1.04\times$ to $4.2\times$, with the largest overhead on sim-to-real mug cleanup at approximately 180,000 observation windows. Additional cost comes mainly from the no-gradient passes used to extract features and evaluate per-sample losses, together with data loading. Multi-source weight refresh takes $1.1$--$1.4\times$ the single-source refresh time.